\pdfoutput=1

\documentclass[11pt]{article}

\usepackage[]{acl}

\usepackage{times}
\usepackage{latexsym}

\usepackage[T1]{fontenc}

\usepackage[utf8]{inputenc}

\usepackage{microtype}

\usepackage{inconsolata}

\usepackage{tabularx}
\usepackage{booktabs}
\usepackage{graphicx}
\usepackage[position=b]{subcaption}
\usepackage{soul} 
\usepackage{lipsum}  
\usepackage{todonotes}
\usepackage[inline]{enumitem}
\usepackage{float}
\usepackage{mathabx} 

\usepackage{listings} 
\usepackage{xcolor} 
\usepackage{multirow}

\definecolor{backcolour}{rgb}{0.95,0.95,0.92}

\lstdefinestyle{mystyle}{
    backgroundcolor=\color{backcolour},   
    basicstyle=\footnotesize\ttfamily,
    breakatwhitespace=false,         
    breaklines=true,                 
    captionpos=b,                    
    keepspaces=true,                 
    numbers=left,                    
    numbersep=5pt,                  
    showspaces=false,                
    showstringspaces=false,
    showtabs=false,                  
    tabsize=2
}

\usepackage{colortbl}

\newcommand{\lightmidrule}{%
    \addlinespace
    \midrule%
    \addlinespace
}

\newcommand{\lighttoprule}{%
    \midrule%
    \addlinespace
}

\newcommand{\lightbottomrule}{%
    \addlinespace
    \midrule%
}

\arrayrulecolor{gray!50}

\title{
Turning English-centric LLMs Into Polyglots: \\
How Much Multilinguality Is Needed?
}

\author{Tannon Kew$^{1}$\thanks{\hspace{0.2cm}Work partially conducted during a research internship at Textshuttle.} \hspace{0.6cm} Florian Schottmann$^{2,3}$ \hspace{0.5cm}  Rico Sennrich$^{1,4}$ \medskip \\
  $^1$University of Zurich,
  $^2$Textshuttle,
  $^3$ETH Zurich,
  $^4$University of Edinburgh \medskip \\
  \texttt{\{kew,sennrich\}@cl.uzh.ch, schottmann@textshuttle.com}}

\begin{document}
\maketitle
\begin{abstract}

The vast majority of today's large language models (LLMs) are English-centric, having been pretrained predominantly on English text.
Yet, in order to meet user expectations, models need to be able to respond appropriately in multiple languages once deployed in downstream applications. This requires strong cross-lingual transfer abilities.
In this work, we investigate the minimal amount of multilinguality required during finetuning to elicit cross-lingual generalisation in English-centric LLMs.
In experiments across four LLMs, we find that multilingual instruction tuning with as few as two to three languages is both necessary and sufficient to elicit effective cross-lingual generalisation, with the limiting factor being the degree to which a target language is seen during pretraining.
Evaluations on five different tasks further reveal that multilingual instruction tuning is most beneficial for generative tasks that assume input/output language agreement, such as in chat settings, while being of less importance for highly structured classification-style tasks.\footnote{Our code and data is available at \url{https://github.com/ZurichNLP/multilingual-instruction-tuning}.}

\end{abstract}

\section{Introduction}

Conversational instruction tuning is a popular method for aligning large pretrained language models (LLMs) with user expectations such that they can effectively respond to a user's input query and follow natural language instructions \cite{ouyang2022training, wang-etal-2023-self-instruct, vicuna2023, zhou2023lima}.
An implicit expectation of conversational chatbots is that the language of a model's response should match that of the user's input query.
For instance, unless otherwise specified, a German-language input should result in a German-language output.
However, since the vast majority of training data is in English, many instruction-tuned LLMs struggle to respond consistently in other languages \cite{ouyang2022training, touvron2023llama2, chen-etal-2024-monolingual, ye2023language, zhang-etal-2024-plug}.

\begin{figure}[t!]
     \centering
     \includegraphics[width=\linewidth]{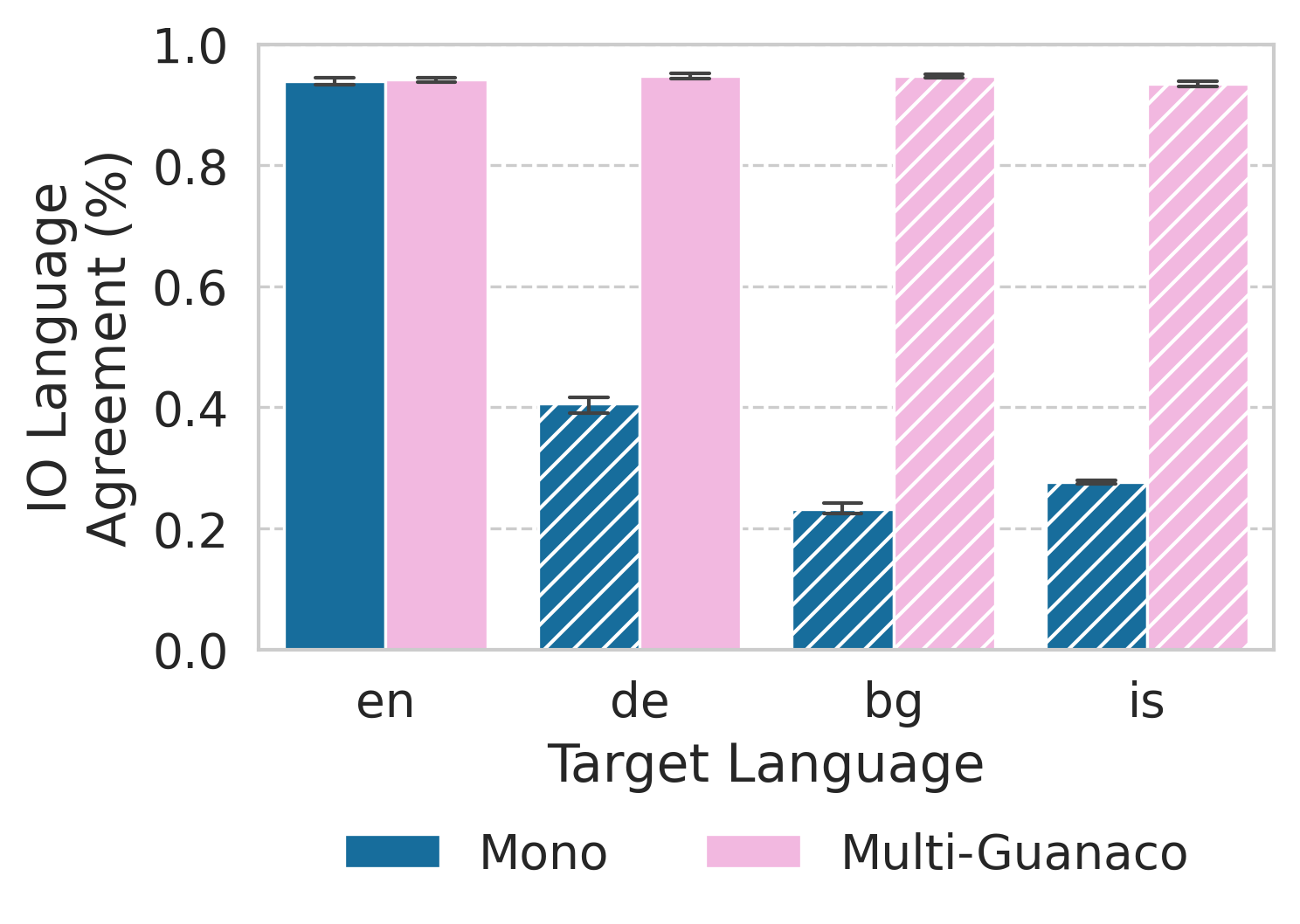}
    \caption{
    Input/output (IO) language agreement for English (en), German (de), Bulgarian (bg), and Icelandic (is) given English-only instruction tuningono) or andmultilingual instruction tuning (Multi-Guanaco).
    Striped bars indicate that the target language is not seen during finetuning (i.e.\ the 0-shot setting). 
    Error bars show a confidence interval of 95\%.
    }
\label{fig:chat_lang_match_llama2_7b_mono_vs_guanaco}
\end{figure}

Despite limited exposure, however, English-centric LLMs such as Llama 2 seemingly achieve near-perfect input/output (IO) language agreement when tuned with relatively few multilingual conversational instructions. 
Figure \ref{fig:chat_lang_match_llama2_7b_mono_vs_guanaco} depicts IO language agreement, as measured by OpenLID \cite{burchell-etal-2023-open}, and compares multilingual tuning on the language-diverse Guanaco dataset \cite{dettmers2023qlora, kopf2023openassistant} with monolingual tuning on an English-only subset of instructions.
As can be seen, multilingual tuning elicits strong IO language agreement -- within the bounds of OpenLID's error rates -- for non-English languages seen to varying degrees during pretraining and finetuning without degrading performance on English.

This observation raises two major questions, which we aim to address in this paper:
\linebreak
\textbf{Q1}: How much multilinguality is required during finetuning to elicit 0-shot cross-lingual generalisation in English-centric LLMs?
\linebreak
\textbf{Q2}: Which languages and tasks benefit most from multilingual instruction tuning of English-centric LLMs?

To investigate these questions, we instruction tune four distinct English-centric LLMs with varying degrees of multilinguality and evaluate performance on a diverse set of up to 19 languages across five different tasks.
Specifically, we consider high-, medium-, and low-resource target languages with regard to a model's pretraining data distribution and generative tasks, such as single-turn dialogue, sentence simplification, and extractive question answering, as well as more structured tasks aimed at assessing commonsense reasoning and language understanding. 

Our results indicate that multilingual instruction tuning is crucial for eliciting cross-lingual transfer on generative tasks that assume IO language agreement while being less important in structured tasks that are commonly used to benchmark LLM performance.
Furthermore, we empirically show that only a small number of finetuning languages is required to promote cross-lingual transfer.
This highlights that tuning data for all potential target languages is not necessary to derive capable polyglot chat models from English-centric LLMs.

\section{Related Work}

\subsection{Instruction Tuning LLMs}

Unlike task-specific finetuning, instruction tuning aims to promote cross-task generalisation, allowing for a `generalist' model that is capable of completing any text-based task on the basis of natural language instructions provided at inference time \cite{mishra2022crosstask, wei2022finetuned, wang-etal-2022-super, sanh2022multitask, longpre2023flan}.
Meanwhile, framing instructions in a conversational manner and over multiple dialogue turns has been shown to be effective at deriving performant chat models \cite{taori2023alpaca, dolly2023, vicuna2023, dettmers2023qlora, ding2023enhancing}.
Furthermore, LLM instruction tuning remains effective given relatively limited labelled data \cite{ouyang2022training, touvron2023llama2, zhou2023lima}, parameter efficient training strategies \cite{hu2021lora, zhang2023llamaadapter} and model quantisation \cite{dettmers2023qlora, li2023loftq}.

\subsection{Cross-lingual Transfer in English-centric LLMs}

The vast majority of today's publicly available LLMs are English-centric.
For instance, GPT-3's training data consisted of approximately 93\% English documents with the remaining 7\% pertaining to other languages \cite{brown2020}.\footnote{\url{https://github.com/openai/gpt-3/blob/master/dataset_statistics}}
This trend is further reflected in popular open-weight LLMs (see Table \ref{tab:language_distributions}).
One potential reason for this could be the ``the curse of multilinguality'' \cite{conneau-etal-2020-unsupervised} which arises from having to share a finite model capacity across more languages \cite{lin-etal-2022-shot, le-scao-etal-2022-language}.
Consequently, cross-lingual transfer abilities of performant English-centric models is highly valuable.

Despite this, proprietary models such as GPT-3 and its derivatives have shown impressive performance in multilingual settings across a range of NLU and NLG benchmarks \cite{lai2023chatgpt, holmstrom-etal-2023-bridging, armengol-estape-etal-2022-multilingual, hendy2023good, lu2023chainofdictionary, jiao2023chatgpt, bang2023multitask, laskar-etal-2023-systematic}. 
Meanwhile, there is a growing body of research dedicated to studying similar capabilities of open-weight LLMs \cite{bawden-yvon-2023-investigating} and the benefits of multilingual tuning \cite{ye2023language, muennighoff-etal-2023-crosslingual}. 
For instance, large-scale multilingual instruction tuning has been shown to improve performance on open-ended chat tasks in multiple target languages for both English-centric and multilingual LLMs \cite{chen-etal-2024-monolingual, li2023bactrianx, weber2024investigating}.
In concurrent work to ours, \citet{shaham-etal-2024-multilingual} report that this holds given minimal multilingual instruction tuning with the multilingual PaLM 2 model \cite{anil2023palm2technicalreport}.

Our work closely relates to that of \citet{chen-etal-2024-monolingual} and \citet{shaham-etal-2024-multilingual} insofar as we explore how multilingual instruction tuning can improve cross-lingual transfer in LLMs.
However, in contrast to these works, we focus on the \textit{minimal amount of multilinguality} needed to elicit cross-lingual transfer in \textit{open-weight English-centric LLMs}, and we evaluate on a range of tasks and target languages that cover the full distribution of a model's pretraining data.

\section{Experimental Setup}
\label{sec:experimental_setup}

To explore the multilingual capabilities of English-centric LLMs, we instruction tune a series of models on a fixed-size set of examples, varying the number of languages available.
Following this, we evaluate the resulting models in multiple target languages on five distinct tasks that are representative of how LLMs may be used in downstream applications.

\subsection{English-centric LLMs}

A prevailing trend in the development of recent LLMs is a clear focus on scaling up the size of the pretraining corpus \cite{hoffmann2022training, sardana2023chinchillaoptimal}. 
For instance, open-weight LLMs such as Falcon \cite{almazrouei2023falcon} and Llama 2 \cite{touvron2023llama2} were pretrained on 1.5 and 2 trillion tokens, respectively.
Yet, while this far surpasses the 300 billion tokens used to train GPT-3 \cite{brown2020}, the distribution of language coverage remains similar with more than 90\% pertaining to English (see Table \ref{tab:language_distributions}).
For our main experiments, we focus on Llama 2 7b. 
We then consider Llama 2 70b (\S\ref{sec:scaling_experiments}) to study the effect of model scaling on multilinguality.
Additionally, we also test whether our findings generalise to other LLMs using Falcon 7b and Llama 3 8b \cite{metaai_introducing_2024}, which employ different pretraining approaches (see Appendices \ref{app:falcon_results} and \ref{app:llama3_results}).

\subsection{Instruction-tuning Data}
\label{sec:finetuning_data}

For instruction tuning, we take inspiration from \citet{dettmers2023qlora} and finetune on high-quality conversations from the OpenAssistant dataset \cite{kopf2023openassistant}.
These conversations comprise multiple dialogue turns between crowdworkers who were asked to either interact with or assume the role of a helpful AI assistant.
In contrast to \citet{dettmers2023qlora}, who use all 9,846 top-rated conversations to train their `Guanaco' models, we subsample training instances from the Guanaco dataset in order to control the amount of multilinguality.
Specifically, we sample 3,200 unique English instances as an initial monolingual dataset, which we refer to as `Mono'.
To construct datasets for multilingual finetuning we sample 200 unique training examples from each of the five most frequent non-English languages in Guanaco (Spanish, Russian, German, Chinese, and French).
Given these subsets, we incrementally substitute English examples in Mono for non-English ones, one language at a time, following the order of how frequently each language appears in Guanaco. 
The resulting multilingual datasets are denoted as Multi-$i$, where $i$ indicates the number of distinct languages included.
Table \ref{tab:multi_i_dataset_distributions} provides a detailed summary of the makeup of these datasets.
For comparison, we also train models on the full Guanaco dataset, which includes more than 30 distinct languages.

\subsection{Instruction-tuning and Inference Settings}

For instruction tuning, we train LoRA adapters ($R=64$, $\alpha=16$) \cite{hu2021lora} using Hugging Face's TRL library\footnote{\url{https://github.com/huggingface/trl}}. 
All models are trained for 2k update steps on sequences of 1024 tokens using an effective batch size of 64 and a learning rate of $1\mathrm{e}{-5}$.\footnote{Details on hardware are provided in Appendix \ref{sec:training_details}.}

For efficient inference, we employ vLLM \cite{kwon2023efficient}.
We use nucleus sampling \cite{holtzman2019curious} ($p=0.9$) with a temperature of 0.8 for open-ended generation tasks and a temperature of 0.001 for the more constrained QA-style tasks to encourage more deterministic behaviour.
For all tasks, we report the mean performance across three inference runs using different random seeds.

\subsection{Evaluation Tasks and Languages}

As evaluation tasks, we consider single-turn dialogue, sentence simplification, extractive question answering, commonsense reasoning and natural language inference.
Since all of these tasks differ in terms of the availability and representation of ground-truth labels, we describe the specific evaluation strategies in the following section.

\begin{figure*}[ht!]
    \centering
    \includegraphics[width=\textwidth]{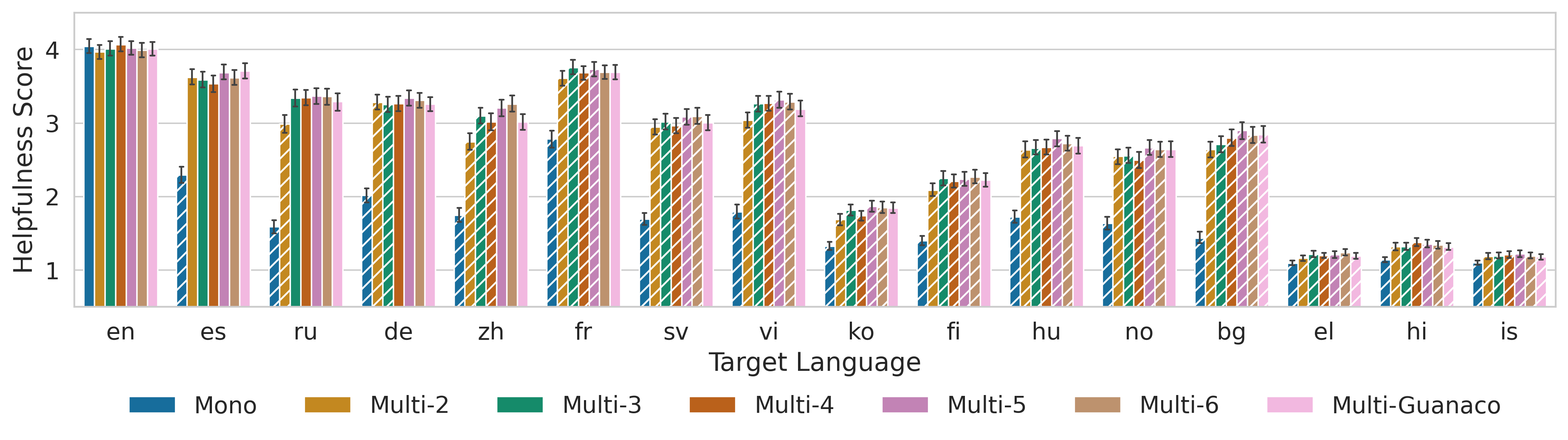}
    \caption{
    Average helpfulness of single-turn dialogue responses from Llama 2 7b given incremental multilingual instruction tuning.
    Striped bars indicate a 0-shot setting and error bars show a confidence interval of 95\%.
    }
    \label{fig:chat_llm_judge_llama_7b_incremental}
\end{figure*}

We select target languages based on the makeup of Llama 2's pretraining data (see Table \ref{tab:language_distributions}).
In doing so, we aim to study the effect of multilingual instruction tuning in both supervised and 0-shot cross-lingual settings \cite{wu-dredze-2019-beto},
and report results on high-, medium-, and low-resource languages with regard to the pretrained LLM, covering distinct language families and scripts.
German (de), French (fr), Swedish (sv), Chinese (zh),\footnote{Note, \citet{touvron2023llama2} do not distinguish between Simplified and Traditional Chinese in their reporting. For our purposes, we explicitly use Simplified Chinese.} Spanish (es), Russian (ru), and Italian (it) represent \textbf{high-resource} languages with an estimated 3.4 to 2.2 billion tokens.
Portuguese (pt), Vietnamese (vi), Korean (ko), Finnish (fi), Hungarian (hu), Norwegian (no),\footnote{Again, lacking specification between Norwegian Bokmål and Nynorsk in \citet{touvron2023llama2}, we explicitly consider the former.} Bulgarian (bg), and Slovenian (sl) represent \textbf{medium-resource} languages, with an estimated 1.8 to 0.2 billion tokens.
And finally, we select Greek (el), Hindi (hi), and Icelandic (is) as \textbf{low-resource} languages, whose frequency in Llama 2's pretraining data is not known but are likely appear in small amounts due to contamination \cite{blevins-zettlemoyer-2022-language}.
Unfortunately, existing benchmark evaluation datasets do not cover all of these languages and thus we limit the target languages in those tasks to the available subset of our target languages.

\section{Experiments}
\label{sec:experiments}

\subsection{Single-turn Dialogue}
\label{sec:chat}

General-purpose chatbots are a popular application of instruction-tuned LLMs.
To evaluate performance in this type of setting, we make use of the AlpacaEval prompt dataset \cite{dubois2023alpacafarm},
which includes a diverse set of prompts for open-ended questions, creative writing, brainstorming, and other tasks.
We randomly sample 300 prompts and translate these into each target language. 
As a translation engine, we follow \citet{lai2023okapi} and use GPT-3.5-Turbo.\footnote{To translate prompts from AlpacaEval, we use \texttt{gpt-3.5-turbo-0613}.}
In contrast to using dedicated translation systems, employing GPT-3.5-Turbo for this purpose has the advantage of being able to explicitly specify instructions that allow for preserving code blocks, tables, and terminology, which we include as part of our translation prompt (see Figure \ref{app:translating_alpacaeval}).
Furthermore, since GPT-3.5-Turbo is trained on instruction- and conversational-style data, we expect it to perform well at translating in this domain.

To automatically evaluate open-ended responses, we leverage LLM-as-a-judge \cite{zheng_judging_2023}. 
Following \cite{zhou2023lima}, given an input prompt and model's response, we ask GPT-3.5-Turbo\footnote{For evaluation, we use \texttt{gpt-3.5-turbo-1106} due to its longer context window.} to grade the \textit{helpfulness} of the response on a 6-point Likert scale (see Figure \ref{fig:eval_helpfulness_prompt} for the prompt used).
For each evaluation instance, we provide the prompt and model-generated response \textit{directly} in the target language, which we found to be on par with evaluating via first translating responses into English \cite[cf.][see Appendix \ref{app:direct_vs_translated_eval} for more details]{hada-etal-2024-large}.
As noted by \citet{chen-etal-2024-monolingual}, GPT-3.5-Turbo sometimes ignores the fact that the output language differs from the input language. Therefore, we force a score of 1 (indicating least helpful) if the language of the response does not match the intended target language according to OpenLID \cite{burchell-etal-2023-open}.

\paragraph{Results}

\begin{figure*}[ht!]
    \centering
    \includegraphics[width=\textwidth]{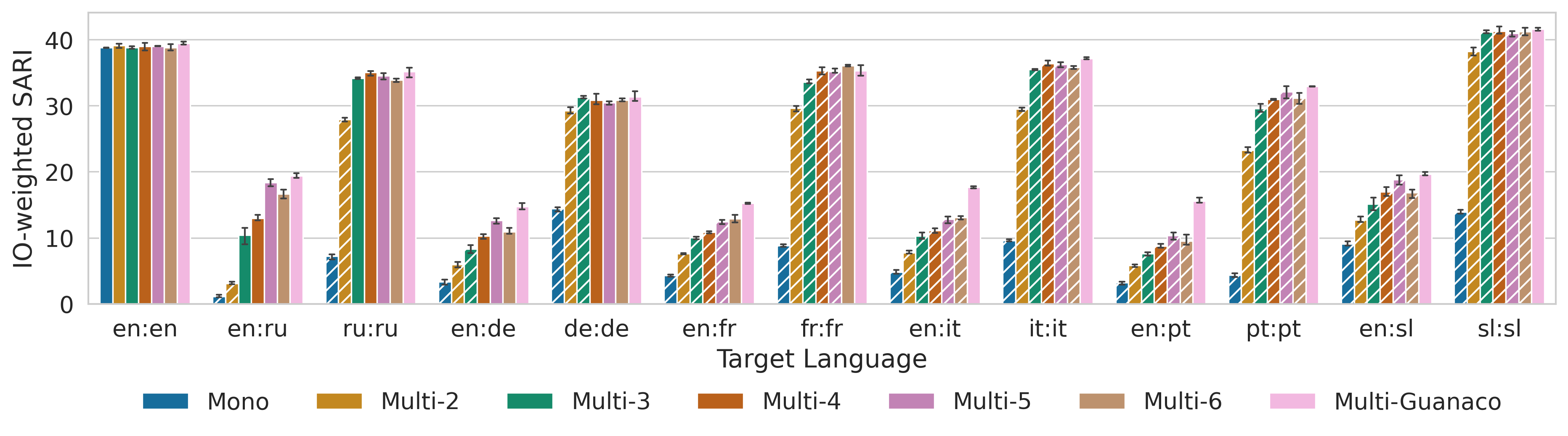}
    \caption{
    SARI weighted by IO language agreement for Llama 2 7b given incremental multilingual instruction tuning. 
    Results are shown for both cross-lingual prompting (en:xx) and monolingual prompting (xx:xx).
    Striped bars indicate a 0-shot setting and error bars show a confidence interval of 95\%.
    }
    \label{fig:multisim_llama_7b_incremental}
\end{figure*}

Figure \ref{fig:chat_llm_judge_llama_7b_incremental} shows the helpfulness scores assigned by the LLM judge on 16 target languages.
For English, performance remains uniform across all multilingual instruction tuning settings.
For high- and medium-resource non-English target languages, performance increases significantly when moving from monolingual to bilingual instruction tuning (Multi-2).
Meanwhile, performance plateaus when training with as few as three languages (Multi-3), with no substantial differences observed between supervised and 0-shot settings.
For low-resource languages, performance remains low despite multilingual finetuning.
Manual inspection reveals these outputs, while initially convincing and sufficient for language identification (see Figure \ref{fig:chat_lang_match_llama2_7b_mono_vs_guanaco}), are mostly nonsensical.
These results indicate that instruction tuning with as few as two to three languages is necessary and sufficient to elicit 0-shot cross-lingual generalisation among high- and medium-resource languages on this task, with language exposure during pretraining being the main limitation to this generalisation ability.

\subsection{Sentence Simplification}
\label{sec:ts}

Sentence simplification aims to make complex sentences easier to read and understand.
Due to a lack of high-quality supervised training data, automatic sentence simplification remains a challenging task \cite{stajner-2021-automatic}, and thus stands to benefit from the few- and 0-shot generalisation capabilities of instruction-following LLMs.
To assess performance on this task, we use the MultiSim benchmark \cite{ryan-etal-2023-revisiting}, which includes sentence-aligned parallel datasets in multiple languages.
Since the individual datasets in MultiSim are taken from distinct sources, the amount of available data varies across languages.
We sample 1,371 complex-simple sentence pairs for each language, except for Slovenian, where we use all 939 instances available (see Appendix \ref{app:multisim} for details).

To prompt models for sentence simplification, we repurpose the detailed instructions given to crowdworkers for the creation of the ASSET corpus \cite{alva-manchego-etal-2020-asset}, which has been shown to be effective in few-shot settings \cite{kew-etal-2023-bless}.
In its original form, this prompt explicitly states that the output should be suitable for ``learners of English''. 
We remove this language specification so that the model is not explicitly instructed on the target language and instead is expected to infer it from the complex input sentence provided.
To translate the prompt into each of our considered target languages, we used a free-tier machine translation service.\footnote{\url{https://www.deepl.com/translator}}
We also consider both a cross-lingual prompt setting (en:xx) and a monolingual prompt setting (xx:xx).
The former presents the task instruction in English and provides the complex source sentence in the relevant target language, while the latter is entirely in the target language.
An example of the prompt template is shown in Table~\ref{tab:ts_prompt}.

For evaluation, we report SARI \cite{xu-etal-2016-optimizing} as an indicator of simplicity and adequacy given the reference simplification.
To account for IO language agreement, we weight a model's corpus-level SARI scores by the percentage of IO language agreement it achieved.
This penalises model if it consistently produces outputs in a language that does not match the source sentence.
For example, if the model generates English outputs given German input sentences at a rate of 50\%, the final score is reduced by half.

\begin{figure*}[ht!]
    \centering
    \includegraphics[width=\textwidth]{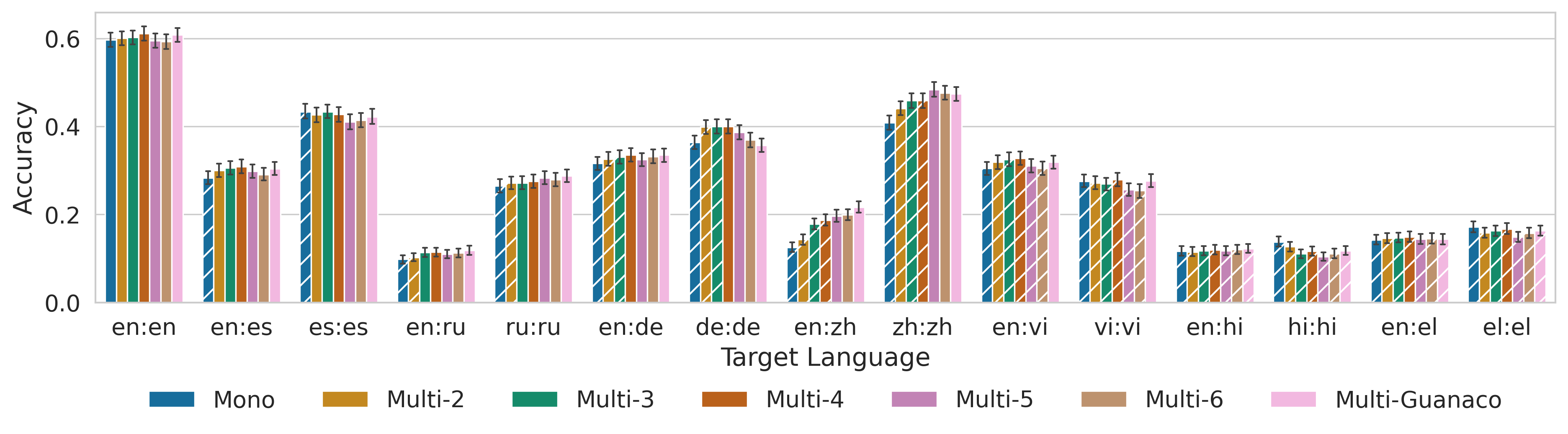}
    \caption{
    XQuAD results for Llama 2 7b given incremental multilingual instruction tuning.
    Results are shown for both cross-lingual prompting (en:xx) and monolingual prompting (xx:xx). 
    Striped bars indicate a 0-shot setting and error bars show a confidence interval of 95\%.
    }
    \label{fig:xquad_llama_7b_incremental}
\end{figure*}

\paragraph{Results}

Figure~\ref{fig:multisim_llama_7b_incremental} shows the performance according to IO-weighted SARI on each of the target languages considered given incremental multilingual instruction tuning.
Similar to the results on the single-turn dialogue task, performance remains uniform for English as the ratio of multilingual instructions increases.
For all non-English target languages, we observe a large discrepancy between the cross-lingual prompting strategy and the monolingual prompting strategy.
While the former tends to improve as the number of languages increases, it fails to match that of the monolingual prompting strategy under all conditions.
In contrast, the results for the monolingual prompting strategy closely resemble those of the single-turn dialogue task across all available target languages: performance gains are most pronounced when moving from monolingual to bilingual instruction tuning, and they generally plateau with as few as three instruction tuning languages.

\subsection{Extractive Question Answering}
\label{sec:xquad}

In contrast to open-ended questions commonly used to query LLMs in chat settings, extractive question answering requires the model to identify relevant answer spans within longer context passages provided as part of the prompt.
Such a task closely resembles a retrieval augmented generation (RAG) setting, which is a popular method for extending an LLM's knowledge with additional data not available during training \cite{lewis2020retrieval, izacard-grave-2021-leveraging}. 
To evaluate our models on this task in multiple target languages we use XQuAD \cite{artetxe-etal-2020-cross}.\footnote{We report results measured on the validation split of XQuAD since labels for the test split are not publicly available. This provides 1,190 QA pairs that were professionally translated into different languages.}

As a starting point, we borrow the English prompt from \citet{lai2023chatgpt} and manually translate it into each of the target languages considered. 
Additionally, we include a standardised response prefix as part of the prompt, effectively force-decoding the response ``Based on the passage, the answer to the question is''. 
This allows us to better isolate the relevant answer string in the generative model's output. 
A response is considered correct if the ground truth answer can be matched with the beginning of the model's response after minimal post-processing.\footnote{We find that some post-processing of model outputs is necessary for certain languages. Specifically, when queried with German and Russian prompts, models consistently repeated the question before providing the extracted answer. To handle such cases, we strip away the repeated question and truncate the response to a maximum of 50 characters or the first line break, whichever comes first.}
Again, we consider both cross-lingual (en:xx) and monolingual (xx:xx) prompting strategies.
An example of our prompting strategy for this task and model outputs is shown in Table \ref{tab:xquad_example}.

\paragraph{Results}

Figure \ref{fig:xquad_llama_7b_incremental} shows the performance on each of our target languages in XQuAD given incremental multilingual instruction tuning.
Again, we observe consistent performance for English under multilingual instruction tuning settings. 
Meanwhile, performance on high-resource languages such as German and Chinese exhibits marginal gains when moving from monolingual to bilingual instruction tuning and plateaus with relatively few languages.
For low-resource languages (e.g., Hindi and Greek), performance remains consistently low across all settings, indicating that the model's ability to generalise to these languages is limited by the lack of exposure during pretraining.
Comparing the cross-lingual and monolingual prompting strategies, we observe that the latter generally leads to better performance. 
This agrees with the results from the sentence simplification task (\S\ref{sec:ts}) and highlights the benefit of prompting under an IO language agreement assumption.
While performance gains on this task are generally less pronounced than on the previous tasks considered, we note that 0-shot extractive QA is inherently challenging for LLMs tuned on conversational instructions as they tend to generate verbose responses rather than the single word or entity that correctly matches the ground truth.

\subsection{Commonsense Reasoning}
\label{sec:xcsqa}

\begin{figure*}[ht]
    \centering
    \includegraphics[width=\textwidth]{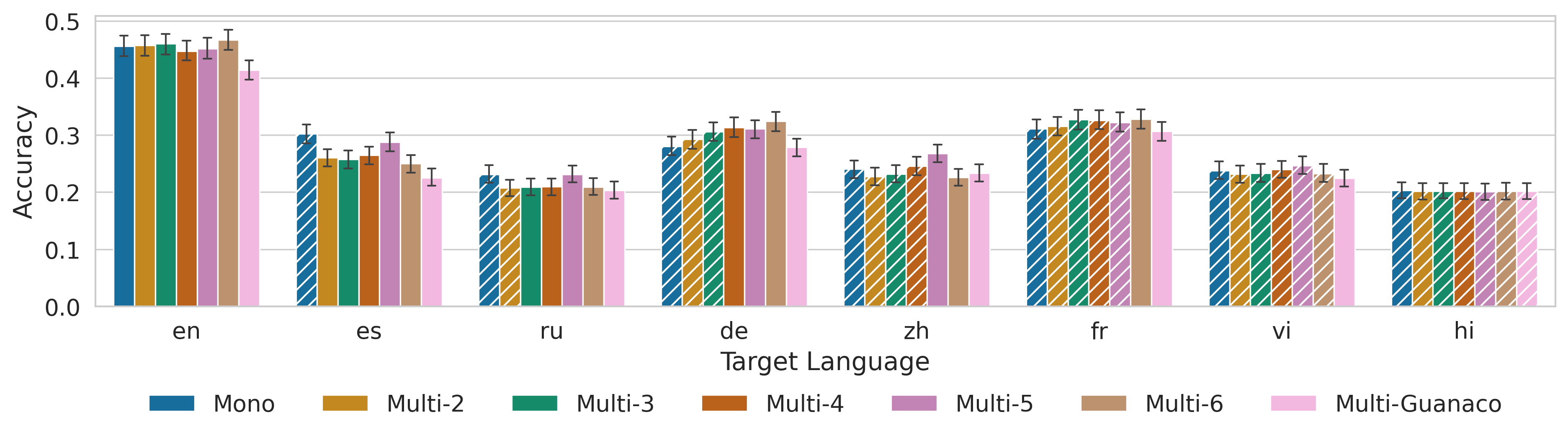}
    \caption{
    X-CSQA results for Llama 2 7b given incremental multilingual instruction tuning.
    Striped bars indicate a 0-shot setting and error bars show a confidence interval of 95\%.
    }
    \label{fig:xcsqa_llama_7b_incremental}
\end{figure*}

\begin{figure*}[ht!]
    \centering
    \includegraphics[width=\textwidth]{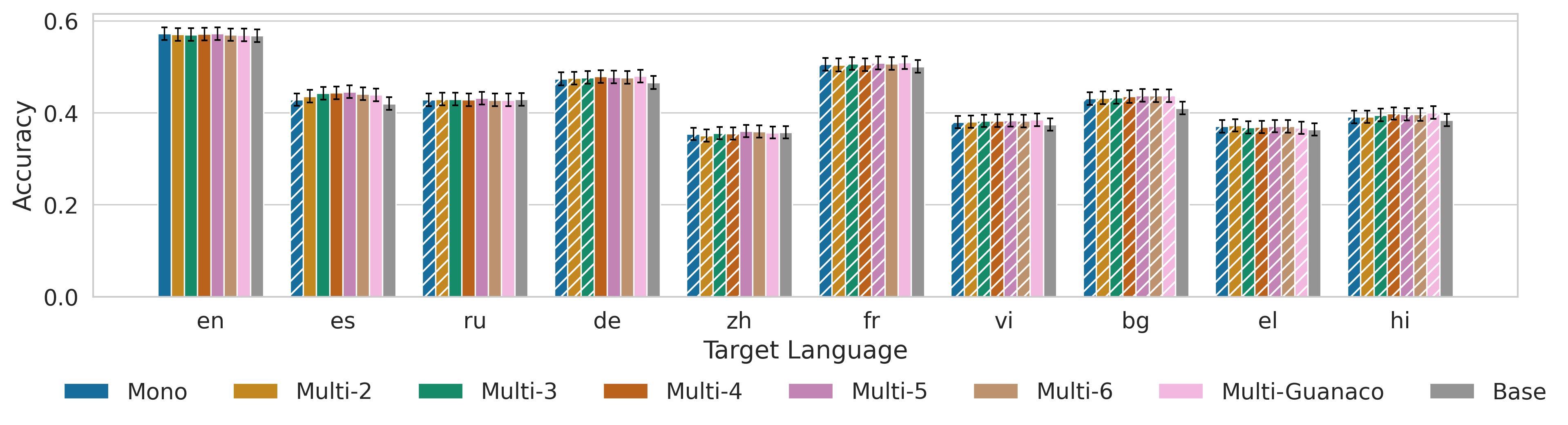}
    \caption{
    XNLI results for Llama 2 7b given incremental multilingual instruction tuning.
    Striped bars indicate a 0-shot setting and error bars show a confidence interval of 95\%.
    }
    \label{fig:xnli_llama2_7b_incremental}
\end{figure*}

`Commonsense knowledge' is the term frequently used to refer to the set of general facts that reflect how concepts can relate to one another in the real world \cite{li-etal-2022-systematic, talmor-etal-2019-commonsenseqa}.
Effectively understanding natural language requires some representation of these concepts and relationships, making commonsense reasoning a key skill for LLMs.
To evaluate how well English-centric models can apply commonsense reasoning across languages, we use the X-CSQA dataset \cite{lin-etal-2021-common}.\footnote{We report results for X-CSQA measured on the validation set of 1,000 questions.}
This dataset contains questions paired with multiple choice answers which aim at assessing general, language-agnostic world knowledge involving different types of commonsense reasoning.
Given a question and a set of five possible answers from A-E, we prompt the model to output the letter corresponding to the most suitable answer.
Again, we borrow the English prompt template for this task from \citet{lai2023chatgpt} as a starting point and translate it into each target language. 
For this task we consider only the monolingual prompting strategy (xx:xx), which proved most effective in the previous tasks.
An example of the prompt used is given in Table \ref{tab:xcsqa_example}.

\paragraph{Results}

Figure \ref{fig:xcsqa_llama_7b_incremental} shows the accuracy on X-CSQA given incremental multilingual instruction tuning.
Again, for English, we observe that multilingual instruction tuning does not significantly degrade performance.
However, in contrast to the previous tasks considered, we do not observe any consistent gains for non-English target languages on the basis of multilingual instruction tuning, suggesting that there are limitations on what tasks this benefits.

\begin{figure*}[ht]
    \centering
    \includegraphics[width=\textwidth]{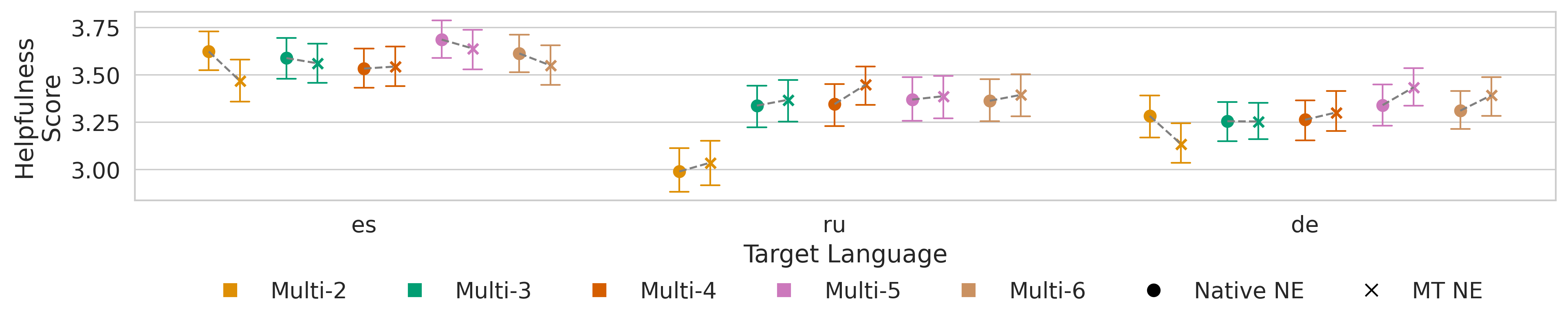}
    \caption{
    Comparison of Llama 2 7b given multilingual instruction tuning using native non-English (Native NE) examples vs.\ translated non-English (MT NE) examples.
    Error bars show a 95\% confidence interval.
    }
    \label{fig:chat_llm_judge_llama_7b_incremental_diff_mt}
\end{figure*}

\subsection{XNLI}
\label{sec:xnli}

Natural language inference (NLI) is an important skill for LMs, as it ensures textual coherence, which is particularly critical as input and output sequences grow in length.
Given two sentences, this task aims to recognise a relationship between them either as \textit{entailment}, \textit{contradiction}, or \textit{neutral}. 
To assess an LLM's ability to solve this task given multilingual instruction tuning, we use XNLI \cite{conneau-etal-2018-xnli} and evaluate performance on the official test split. 

For this task, we use the implementation in the LM Evaluation Harness \cite{gao-etal-2023-eval_harness}.\footnote{\url{https://github.com/EleutherAI/lm-evaluation-harness}}
Instead of querying the model to generate the desired label, we use rank classification \cite{sanh2022multitask, brown2020}, which allows for direct comparison to the base model. 
In this setting, multiple queries are constructed for each test instance given a predefined template and subsequently scored by the model.
The most probable query sequence under the model is then chosen as the final answer.
Prompt templates for this task are language-specific making it equivalent to monolingual prompting (xx:xx).
An example of this task is provided in Table \ref{tab:xnli_example}.

\paragraph{Results}

Figure \ref{fig:xnli_llama2_7b_incremental} shows the accuracy measured on our target languages from XNLI.
Strikingly, performance for all target languages remains uniform regardless of the amount of multilinguality used during finetuning.
Furthermore, the comparable performance across all training settings suggests that, in general, instruction tuning fails to provide measurable gains in this challenging task.

\section{Further Analysis}
\label{sec:additional_experiments}

Our experimental results on five distinct tasks demonstrate that multilingual instruction tuning benefits open-ended single-turn dialogue and sentence simplification in non-English languages most strongly.
In this section, we focus on single-turn dialogue to investigate the impact of instruction diversity vs.\ language diversity and model scaling.

\subsection{Instruction Diversity vs.\ Language Diversity}
\label{sec:instr_vs_lang_diversity}

\begin{figure*}[ht]
    \centering
    \includegraphics[width=\textwidth]{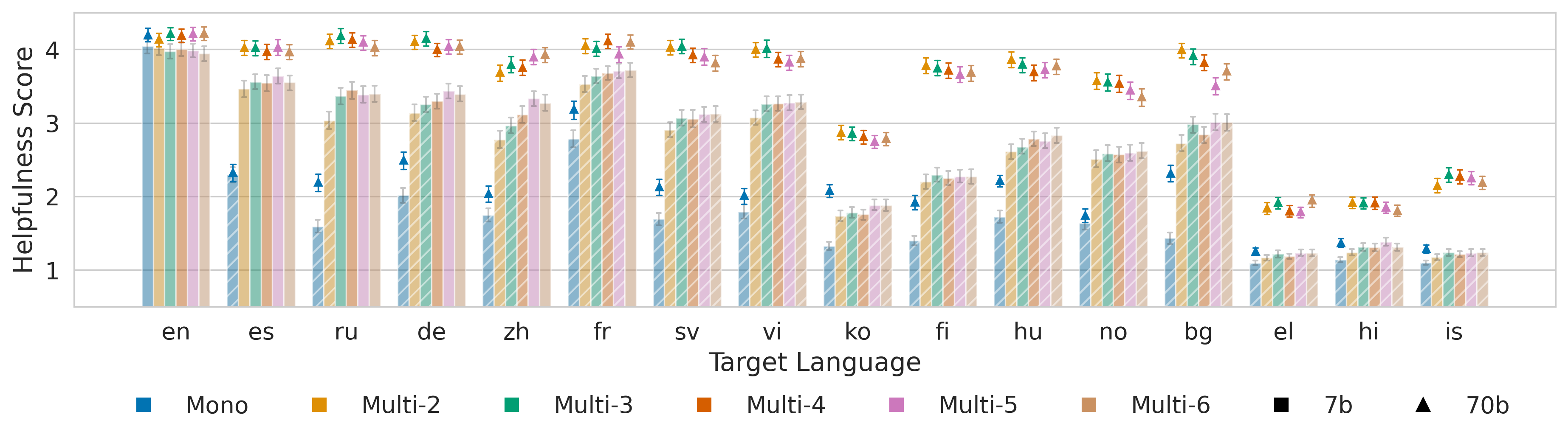}
    \caption{
    Average helpfulness of single-turn dialogue responses from Llama 2 70b (denoted by triangular points) and Llama 2 7b (semi-transparent bars) given incremental multilingual instruction tuning.
    Striped bars indicate a 0-shot setting and error bars show a confidence interval of 95\%.
    }
    \label{fig:chat_llm_judge_llama_7b_incremental_diff_70b}
\end{figure*}

Diversity is a key factor for sample efficient instruction tuning \cite{zhou2023lima}. 
Since our experiments in \S \ref{sec:experiments} make use of native non-English training instances from the OpenAssistant dataset, a potential confounding factor could be that adding more languages during finetuning also introduces more diverse training instructions.
To investigate this, we retrain Multi-$i$ models using translated instruction tuning data from Mono in place of native non-English examples, following the same incremental recipe as described in \S\ref{sec:finetuning_data}.
This ensures that the data distribution remains constant as multilinguality increases.
As a translation engine, we again use GPT-3.5-Turbo\footnote{
Since conversational training instances can be quite long, sometimes exceeding the default 4k token context window, we use \texttt{gpt-3.5-tubo-1106} for this translation task due to its longer context window (16k).
} and the prompt template provided in Figure \ref{fig:ganaco_translation_prompt}.

Figure \ref{fig:chat_llm_judge_llama_7b_incremental_diff_mt} compares the resulting performance on the single-turn dialogue task for a subset of our target languages (results for the remaining target languages are provided in Figure \ref{fig:chat_llm_judge_llama_7b_incremental_diff_mt_additional_langs}).
Notably, we observe no significant differences between tuning with distinctly native non-English examples compared to those derived via automatic translation, indicating that gains attributed to increased multilinguality are not conflated with an increase in the diversity of instructions.
Following this, in Appendix \ref{sec:ablation_experiments}, we dive deeper to investigate the role of language diversity given a fixed budget of non-English training examples.
From these ablation experiments, we find that the impact of increasing the number of languages is stronger than that of increasing the number of examples for a given language.

\subsection{Scaling up Model Size}
\label{sec:scaling_experiments}

In order to investigate the effect of model scaling, we repeat our single-turn dialogue experiments using Llama 2 70b as the underlying base model.
Figure \ref{fig:chat_llm_judge_llama_7b_incremental_diff_70b} shows the resulting helpfulness scores assigned by the LLM judge, with a direct comparison to Llama 2 7b.
Most notably, performance on high- and medium-resource non-English target languages is dramatically improved, often matching that of English.
However, despite these gains, multilingual instruction tuning with two languages remains essential to elicit cross-lingual generalisation on this task, further underpinning our main findings.
In addition, we observe that the larger model's performance on most non-English target languages tends to plateau with just two finetuning languages, unlike the smaller model that typically required three.
Finally, while the performance on low-resource languages remains poor, it exhibits a substantial relative improvement compared to the 7-billion-parameter model.
These results indicate that model scaling is extremely beneficial for exploiting the multilingual capabilities in English-centric models and aligns with findings from \citet{shaham-etal-2023-causes}.

\section{Discussion}
\label{sec:discussion}

Our findings show that multilingual instruction tuning aids cross-lingual transfer in English-centric LLMs, though its effectiveness on downstream performance varies across tasks and the degree to which a specific target language is seen during pretraining.
For high- and medium-resource languages, we observe substantial performance gains on single-turn dialogue and sentence simplification with monolingual prompting, as well as marginal gains on extractive QA with monolingual prompting.
Meanwhile, for low-resource languages and highly structured tasks that impose a strict constraint on the output space regardless of the input language (e.g., multiple choice QA), multilingual instruction tuning has little impact.
This distinction highlights that multilingual instruction tuning is most beneficial for more open-ended generative tasks that assume IO language agreement, and crucially, that this benefit is limited to high- and medium-resource languages with respect to model's pretraining language distribution.

Most surprisingly, we find that \textbf{multilingual instruction tuning with just two to three languages is necessary and sufficient to promote this generalisation} in all four of the English-centric LLMs considered.
Furthermore, we consistently see that performance tends to plateau rather quickly and adding more instruction tuning languages -- including the target language itself -- typically provides no significant gains. 
This underscores that the performance gains are largely the result of cross-lingual transfer and suggests that there may be diminishing returns associated with scaling up multilingual instruction tuning beyond just a handful of languages.

We posit that our findings align with the \textit{superficial alignment hypothesis} put forward by \citet{zhou2023lima}, which states that a model acquires its knowledge and abilities during pretraining, while instruction tuning simply guides the model towards a desirable `subdistribution of formats' to use when prompted.
We refine this hypothesis by adding that just a small amount of multilingual instruction data also encourages the model to learn a simple mapping between input and output language. 
This guides the model towards \textbf{language-specific subdistributions}, leading to better performance on tasks where IO language agreement is required.

\section{Conclusion}

We investigated the minimal amount of multilingual instruction tuning required to elicit multilingual capabilities of four distinct English-centric LLMs.
Our results show that finetuning with as few as two to three languages is necessary and sufficient to promote effective cross-lingual transfer, allowing models to better exploit the relatively small amounts of non-English data seen during pretraining.
Experiments on five distinct tasks revealed that this can lead to significant performance improvements for high- and medium-resource languages on open-ended generative tasks that assume input/output language agreement.

While the effectiveness of cross-lingual transfer is indeed good news for LLM developers, future work could explore methods to reduce the performance gap between English and non-English languages, particularly for smaller models, as well as investigating tasks for which language-specific instruction tuning may be of greater importance, such as tasks involving cultural awareness.

\section*{Limitations}

Our experimental results demonstrate that a relatively small amount of multilingual instruction tuning data can elicit highly valuable cross-lingual transfer, leading to improved performance in open-ended generation tasks that assume IO language agreement.
However, we have not explored tasks involving more complicated mappings between languages, such as tasks involving extensive cross-lingual processing or translation.

In this work we rely on automatic language identification to construct multilingual training data and in evaluating model outputs. 
To this end, we employed the OpenLID model from \citet{burchell-etal-2023-open}.
Despite low error rates achieved by this model, language identification is not perfect and can lead to some texts being misidentified.
To mitigate the risk of unintentional language contamination \cite{blevins-zettlemoyer-2022-language} in our finetuning datasets we include training examples whose language is identified with a confidence threshold $\geq$ 0.8.

In \S\ref{sec:instr_vs_lang_diversity} we investigate the impact of multilingual diversity versus training example diversity. 
While our findings reveal that there is no significant difference between these two settings, we note that even when finetuning with the original native non-English examples, task diversity may be inherently limited by design of the data collection. 
For instance, regardless of the language used, crowdworkers were asked to follow the same set of guidelines\footnote{\url{https://projects.laion.ai/Open-Assistant/docs/guides/guidelines}} when creating the data.

Finally, to evaluate the quality of model-generated responses in the single-turn dialogue task, we rely on LLM-based evaluation methods.
While a number of works have shown that LLM-based evaluation provides a decent proxy for assessing the quality of dialogue responses, achieving strong correlations with human judgements \cite{liu-etal-2023-g, kocmi-federmann-2023-large, zheng_judging_2023}, we stop short of empirically establishing this agreement on our own model outputs.
To ensure the validity of model outputs we manually assessed the adequacy of randomly sampled responses in three target languages (en, es, and de).
This assessment revealed a similar trend to our LLM-based evaluation, namely that Spanish-language outputs were generally slightly worse than their English-language counterparts but slightly better than those in German.

\section*{Ethical Considerations and Risks}

This work aims to evaluate multilingual capabilities of English-centric LLMs. 
In doing so, we acknowledge potential ethical considerations and risks associated with our research.
Firstly, LLMs have been shown to inadvertently perpetuate biases present in their training data, which can lead to unexpected and unfair outcomes when these models are used in real-world applications. 
Therefore, measures must be taken to minimise this risk (e.g., using additional alignment strategies, rigorous testing in multilingual settings) before deploying public-facing models.
Secondly, when building on top of English-centric LLMs, there is a risk of cultural homogenization, where nuances and diversity of different languages could potentially be lost due to their under-representation in the training data.

\section*{Acknowledgements}

We sincerely thank our friends and colleagues, Fabian Aiolfi, Thea Bächler, Anastassia Shaitarova, Finnur Ágúst Ingimundarson, Cui Ding, Andrianos Michail, Janine Aeberhard, Arnisa Fazla, Farhad Nooralahzadeh, Omnia Ibrahim, Xuan Lam, Manasi Muglikar, Anne Göhring, and Alessia Battisti for helping us with language-specific questions and validating translations used in our experiments, as well as Lena Sophia Bolliger and Patrick Haller for providing valuable feedback on earlier versions of this paper.
Rico Sennrich acknowledges funding by the Swiss National Science Foundation (project MUTAMUR; no.~213976).

\bibliography{anthology_min,custom}

\appendix

\section{Pretraining Data for English-centric LLMs}

With the notable exception of BLOOM \cite{scao2023bloom} and the recent OLMo models \cite{groeneveld-etal-2024-olmo}, very limited information is offered about the distribution of languages represented in datasets used to train LLMs and the subsequent performance on non-English languages.
Table \ref{tab:language_distributions} provides an overview of the document-level language distributions of the LLMs used in this paper.
For Llama 2, the information is taken from the original paper \cite{touvron2023llama2}, in which the authors analyse the training data using a fastText language classifier on corpus documents with a threshold of 0.5.
For GPT-3, we use the official dataset statistics made available on GitHub which provide document-level language identification information.\footnote{\url{https://github.com/openai/gpt-3/blob/master/dataset_statistics/languages_by_document_count.csv}}
To gather statistics for Falcon, we inspected a sample of the RefinedWeb dataset \cite{penedo2023refinedweb} that was constructed to train this family of models.\footnote{\url{https://huggingface.co/datasets/tiiuae/falcon-refinedweb}} 
Using the OpenLID fastText model from \citet{burchell-etal-2023-open}, we identify the most frequent document-level languages based on approximately 320 million examples from this corpus. 
Following \cite{touvron2023llama2}, languages are identified based on a confidence threshold of 0.5 and predictions below this threshold are aggregated under `unknown'.
Note, that while this language distribution has a higher concentration of English data than both Llama 2 and GPT-3, \citet{almazrouei2023falcon} combine the RefinedWeb with additional curated corpora to train the Falcon models, and thus the true language distributions for this model may differ from these estimates.

\begin{table*}[ht!]
    \centering
    \small
    \resizebox{\textwidth}{!}{
        \begin{tabular}{@{}rlr | rlr | rlr@{}}
        \multicolumn{3}{c|}{Llama 2} & \multicolumn{3}{c|}{Falcon} & \multicolumn{3}{c}{GPT-3} \\ 
        \lightmidrule 
        \# & Language & \multicolumn{1}{l|}{Percent} & \# & Language & \multicolumn{1}{l|}{Percent} & \# & Language & \multicolumn{1}{l}{Percent} \\ 
        \lightmidrule
        1 & \textbf{English} & 89.70\% &                1 & \textbf{English} & 95.751\% &                       1 & \textbf{English} & 93.69\% \\
        2 & unknown & 8.38\% &                          2 & unknown & 3.902\% &                                 2 & \textbf{German} & 1.20\% \\
        3 & \textbf{German} & 0.17\% &                  3 & Standard Malay & 0.045\% &                          3 & \textbf{French} & 1.02\% \\
        4 & \textbf{French} & 0.16\% &                  4 & Swahili & 0.038\% &                                 4 & \textbf{Portuguese} & 0.64\% \\
        5 & \textbf{Swedish} & 0.15\% &                 5 & \textbf{French} & 0.019\% &                         5 & \textbf{Italian} & 0.58\% \\
        6 & \textbf{Chinese} & 0.13\% &                 6 & Sardinian & 0.014\% &                               6 & \textbf{Spanish} & 0.51\% \\
        7 & \textbf{Spanish} & 0.13\% &                 7 & \textbf{Portuguese} & 0.014\% &                     7 & Dutch & 0.37\% \\
        8 & \textbf{Russian} & 0.13\% &                 8 & Dutch & 0.01\% &                                    8 & Polish & 0.25\% \\
        9 & Dutch & 0.12\% &                            9 & Chinese (Traditional) & 0.013\% &                   9 & Japanese & 0.25\% \\
        10 & \textbf{Italian} & 0.11\% &                10 & \textbf{Spanish} & 0.010\% &                       10 & Danish & 0.16\% \\
        11 & Japanese & 0.10\% &                        11 & \textbf{Italian} & 0.010\% &                       11 & \textbf{Norwegian} & 0.15\% \\
        12 & Polish & 0.09\% &                          12 & \textbf{German} & 0.009\% &                        12 & Romanian & 0.13\% \\
        13 & \textbf{Portuguese} & 0.09\% &             13 & Danish & 0.008\% &                                 13 & \textbf{Finnish} & 0.13\% \\
        14 & \textbf{Vietnamese} & 0.08\% &             14 & Indonesian & 0.007\% &                             14 & \textbf{Chinese (Simplified)} & 0.12\% \\
        15 & Ukranian & 0.07\% &                        15 & Somali & 0.006\% &                                 15 & \textbf{Russian} & 0.11\% \\
        16 & \textbf{Korean} & 0.06\% &                 16 & \textbf{Swedish} & 0.006\% &                       16 & Czech & 0.10\% \\
        17 & Catalan & 0.04\% &                         17 & \textbf{Russian} & 0.006\% &                       17 & \textbf{Swedish} & 0.06\% \\
        18 & Serbian & 0.04\% &                         18 & Venetian & 0.005\% &                               18 & \textbf{Hungarian} & 0.06\% \\
        19 & Bahasa Indonesian & 0.03\% &               19 & \textbf{Vietnamese} & 0.005\% &                    19 & Chinese (Traditional) & 0.04\% \\
        20 & Czech & 0.03\% &                           20 & Northern Uzbek & 0.005\% &                         20 & Bahasa Indonesian & 0.04\% \\
        21 & \textbf{Finnish} & 0.03\% &                21 & Limburgish & 0.005\% &                             21 & Croation & 0.04\% \\
        22 & \textbf{Hungarian} & 0.03\% &              22 & Tagalog & 0.005\% &                                22 & Turkish & 0.04\% \\
        23 & \textbf{Norwegian} & 0.03\% &              23 & \textbf{Chinese (Simplified)} & 0.005\% &          23 & Catalan & 0.03\% \\
        24 & Romanian & 0.03\% &                        \multicolumn{3}{c|}{\scalebox{0.5}{\vdots}} &           24 & \textbf{Vietnamese} & 0.03\% \\
        25 & \textbf{Bulgarian} & 0.02\% &              27 & \textbf{Norwegian Bokmal} & 0.003\% &              25 & \textbf{Slovenian} & 0.03\% \\
        26 & Danish & 0.02\% &                          \multicolumn{3}{c|}{\scalebox{0.5}{\vdots}} &           26 & Estonian & 0.02\% \\
        27 & \textbf{Slovenian} & 0.01\% &              33 & \textbf{Finnish} & 0.003\% &                       27 & Slovak & 0.02\% \\
        28 & Croatian & 0.01\% &                        \multicolumn{3}{c|}{\scalebox{0.5}{\vdots}} &           28 & \textbf{Korean} & 0.02\% \\
        \multicolumn{3}{c|}{\scalebox{0.5}{\vdots}} &   35 & \textbf{Greek} & 0.002\% &                         \multicolumn{3}{c}{\scalebox{0.7}{\vdots}} \\
        n/a & \textbf{Greek} & n/a &                    \multicolumn{3}{c|}{\scalebox{0.5}{\vdots}} &           31 & \textbf{Greek} & 0.017\% \\
        \multicolumn{3}{c|}{\scalebox{0.5}{\vdots}} &   46 & \textbf{Hungarian} & 0.002\% &                     \multicolumn{3}{c}{\scalebox{0.7}{\vdots}} \\
        n/a & \textbf{Hindi} & n/a &                    \multicolumn{3}{c|}{\scalebox{0.5}{\vdots}} &           45 & \textbf{Hindi} & 0.004\% \\
        \multicolumn{3}{c|}{\scalebox{0.5}{\vdots}} &   51 & \textbf{Slovenian} & 0.001\% &                     \multicolumn{3}{c}{\scalebox{0.7}{\vdots}} \\
        n/a & \textbf{Icelandic} & n/a &                \multicolumn{3}{c|}{\scalebox{0.5}{\vdots}} &           47 & \textbf{Icelandic} & 0.004\% \\
        \multicolumn{3}{c|}{\scalebox{0.5}{\vdots}} &   68 & \textbf{Bulgarian} & 0.001\% &                     \multicolumn{3}{c}{\scalebox{0.7}{\vdots}} \\
        & & &                                           \multicolumn{3}{c|}{\scalebox{0.5}{\vdots}} &            48 & \textbf{Bulgarian} & 0.003\% \\
        & & &                                           78 & \textbf{Icelandic} & 0.001\% &                     \multicolumn{3}{c}{\scalebox{0.7}{\vdots}} \\
        & & &                                           \multicolumn{3}{c|}{\scalebox{0.5}{\vdots}} &           & & \\
        & & &                                           86 & \textbf{Korean} & 0.0003\% &                        & & \\
        & & &                                           \multicolumn{3}{c|}{\scalebox{0.5}{\vdots}} &           & & \\
        & & &                                           102 & \textbf{Hindi} & 0.0001\% &                        & & \\
        \lightbottomrule
        \end{tabular}
    }
    \caption{
    Distribution of document languages in pretraining datasets for popular English-centric LLMs. For Llama 2 and GPT-3, statistics are taken from the original works. For Falcon, statistics are estimated based on a sample of the RefinedWeb corpus (roughly 320 million documents).
    Note that for Llama 2 and Falcon, `unknown' indicates texts identified below a predefined confidence threshold of 0.5. 
    \citet{touvron2023llama2} state that this data partially pertains to programming code.
    Target languages considered in our experiments are highlighted in bold.
    }
    \label{tab:language_distributions}
\end{table*}

\section{Translating Training and Evaluation Data}
\label{app:translation}

As part of our experiments and evaluation, we generate translations of existing datasets. 
All original datasets are licensed under the Apache License 2.0.
Since translations are generated with GPT-3.5-Turbo, OpenAI's usage policy applies to the resulting versions, which we make available for future research.

\subsection{AlpacaEval Prompts}
\label{app:translating_alpacaeval}

\begin{figure}[htb]
\centering
\begin{tikzpicture}[scale=1.0, every node/.style={transform shape}]
\node[rectangle, rounded corners, draw=blue!20, fill=blue!10, text width=0.9\linewidth, align=left, inner sep=1.2ex, font=\ttfamily\small] (prompt) {
You are a helpful assistant.
\linebreak
\linebreak
Translate the following text into \{\{target\_language\}\}. 

Keep the structure of the original text and preserve things like code and names. 

Please ensure that your response contains only the translated text. 

The translation must convey the same meaning as the original and be natural for native speakers with correct grammar and proper word choices. 

Your translation must also use exact terminology to provide accurate information even for the experts in the related fields.
\linebreak
\linebreak
Original: 
\linebreak
\linebreak
\{\{source\_text\}\}
\linebreak
\linebreak
Translation into \{\{target\_language\}\}:
};

\end{tikzpicture}
\caption{
Prompt template used to translate AlpacaEval with GPT-3.5-Turbo. 
Values in curly braces represent placeholders. 
The value specified for `target language' is the English name of the language (e.g., `German') we are translating into, except for Greek and Chinese, where we explicitly specify standard modern Greek and Mandarin Chinese, respectively.
}
\label{fig:translating_alpacaeval_prompt}
\end{figure}

For the evaluation of single-turn dialogue in non-English languages, we translate AlpacaEval prompts from English into each of our considered target languages using GPT-3.5-Turbo and the template in Figure \ref{fig:translating_alpacaeval_prompt}.
Manual inspection of a sampled subset of these translated prompts in various languages revealed that the translations were typically decent, although often included literal translations for metaphorical expressions rather than how a native speaker might express themselves. 
For instance, English `bullet points' was translated literally into Hindi, rather than an arguably more appropriate phrasing such as `important points'. 
Additionally, in languages that distinguish between formal, informal, or gendered pronouns (e.g., German, French, etc.), the formal and male forms are dominant. 
While these characteristics may be not truly representative of how native speakers actually interact with LLMs, we consider these to be potentially valid queries for general-purpose chat models.

\subsection{Guanaco Training Examples}
\label{sec:translating_guanaco}

\begin{figure}[htb]
\centering
\begin{tikzpicture}[scale=1, every node/.style={transform shape}]

\node[rectangle, rounded corners, draw=blue!20, fill=blue!10, text width=0.9\linewidth, align=left, inner sep=1.2ex, font=\ttfamily\small] (prompt) {
You are a helpful assistant.
\linebreak
\linebreak
Translate the following conversation between a human and an AI assistant into \{\{target\_language\}\}. 
\linebreak
Keep the structure of the original text and preserve things like code, names and role labels (e.g. <S1>, <S2>).
\linebreak
Please ensure that your response contains only the translated text. 
\linebreak
The translation must convey the same meaning as the original and be natural for native speakers with correct grammar and proper word choices. 
\linebreak
Your translation must also use exact terminology to provide accurate information even for the experts in the related fields.
\linebreak
\linebreak
Original: 
\linebreak
\linebreak
\{\{training\_instance\}\}
\linebreak
\linebreak
Translation into \{\{target\_language\}\}:
};

\end{tikzpicture}
\caption{
Prompt template used to translate Guanaco training instances from English into various target languages to investigate the effect of language diversity vs.\ instruction diversity.
}
\label{fig:ganaco_translation_prompt}
\end{figure}

To investigate the effect of language diversity compared to instruction diversity (\S \ref{sec:instr_vs_lang_diversity}), we translate a subset of Guanaco's English training examples into non-English target languages and use these to create MT-based Multi-$i$ instruction-tuning datasets.
By default, speaker roles in Guanaco are denoted with `\#\#\# Human:' and `\#\#\# Assistant'. 
To ensure that these are never translated and the dialogue structure is maintained, we substitute them with special tokens `<S1>' and `<S2>' and explicitly tell the model to leave these tokens intact (see Figure \ref{fig:ganaco_translation_prompt}).
Before training, we map the special placeholder tokens back to their original form.

\section{Hardware Requirements for Instruction Tuning}
\label{sec:training_details}

As mentioned in \S \ref{sec:experimental_setup}, we use LoRA \cite{hu2021lora} for parameter-efficient finetuning. 
As hyperparameters, we set $R=64$, $\alpha=16$. 
This results in the number of trainable parameters being roughly 2.3\% of the original model size for Llama 2 7B, 1.2\% for Llama 2 70B and 1.8\% for Falcon 7B.
For 7-billion parameter models, we use two NVIDIA GeForce RTX 3090 with 24GB of memory. 
The time required for each training run is approximately 8 hours.
For the larger 70-billion parameter model, discussed in \S \ref{sec:scaling_experiments}, we use the same hyperparameters and train on four NVIDIA A100 GPUs with 80GB of memory each. 
Here, a single training run takes approximately 20 hours.

\section{MultiSim Datasets}
\label{app:multisim}

MultiSim is composed of 34 distinct text simplification data sets aligned at either the document- or sentence-level \cite{ryan-etal-2023-revisiting}, covering 13 different languages.
Due to restrictive licensing on some datasets, only a subset are openly accessible.
An overview of these datasets is provided in Table \ref{tab:multisim_dataset_statistics}. 
As can be seen, domain coverage and the amount of available data vary considerably across languages.
To balance out the number of evaluation instances between the different languages, we randomly sample up to 1,371 complex-simple sentence pairs from all available corpora for each language, except for Slovenian, where we use all 939 available instances.
While we prioritise sampling from the test and validation splits shared by \citet{ryan-etal-2023-revisiting}, we also draw samples from the designated train splits for Portuguese, German, and Slovenian in order to have a sufficiently large evaluation set for our experiments.

\begin{table*}[ht!]
    \centering
    \small
    \resizebox{0.9\textwidth}{!}{
        \begin{tabular}{clcc}
        \multicolumn{1}{c}{Language} & \multicolumn{1}{c}{Source corpora} & \multicolumn{1}{c}{\# instances avail.\ (test / val / train)} &\multicolumn{1}{c}{\# eval.\ instances} \\
        \cmidrule(lr){1-1} \cmidrule(lr){2-2} \cmidrule(lr){3-3} \cmidrule(lr){4-4} \addlinespace
        \multirow{2}{*}{en} & ASSET \cite{alva-manchego-etal-2020-asset} & 359 / 100 / 19,000 & \multirow{2}{*}{1,371} \\
                            & WikiAuto \cite{jiang-etal-2020-neural} & 5,002 / 4,988 / 576,126 & \\
        \addlinespace
        \addlinespace
        \multirow{4}{*}{ru} & RSSE \cite{sakhovskiy2021rusimplesenteval} & 1,083 / 97 / 3,182 & \multirow{4}{*}{1,371} \\
                            & RuAdapt Encyclopedia \cite{dmitrieva2021quantitative} & 839 / 840 / 7,782 & \\
                            & RuAdapt Fairytale \cite{dmitrieva2021quantitative} & 31 / 31 / 248 & \\
                            & RuWikiLarge \cite{sakhovskiy2021rusimplesenteval} & 312 / 678 / 246,978 &  \\
        \addlinespace
        \addlinespace
        \multirow{2}{*}{de} & GEOLino \cite{mallinson-etal-2020-zero} & 81 / 82 / 958 & \multirow{2}{*}{1,371} \\
                            & TextComplexityDE \cite{naderi2019subjectiveassessmenttextcomplexity} & 26 / 28 / 208 & \\
        \addlinespace
        \addlinespace
        \multirow{2}{*}{fr} & CLEAR \cite{grabar-cardon-2018-clear} & 100 / 294 / 4,196 & \multirow{2}{*}{1,371} \\
                            & WikiLarge FR \cite{cardon-grabar-2020-french} & 345 / 878 / 296,402 & \\
        \addlinespace
        \addlinespace
        \multirow{1}{*}{pt} & PorSimples \cite{aluisio-gasperin-2010-fostering} & 420 / 420 / 6,290 & \multirow{1}{*}{1,371} \\
        \addlinespace
        \addlinespace
        \multirow{5}{*}{it} & AdminIT \cite{miliani-etal-2022-neural} & 49 / 48 / 588 & \multirow{5}{*}{1,371} \\
                            & SIMPITIKI Wiki \cite{tonelli2016simpitiki} & 160 / 146 / 1,436 &  \\
                            & PaCCSS-IT \cite{brunato-etal-2016-paccss} & 1,061 / 1,061 / 60,485 & \\
                            & Teacher \cite{brunato-etal-2015-design} & 17 / 17 / 136 & \\
                            & Terence \cite{brunato-etal-2015-design} & 101 / 102 / 809 & \\
        \addlinespace
        \addlinespace
        \multirow{1}{*}{sl} & SloTS \cite{gorenc2022} & 96 / 94 / 749 & 939 \\
        \lightbottomrule
    \end{tabular}
    }
    \caption{Overview of datasets included in the MultiSim benchmark \cite{ryan-etal-2023-revisiting}.
    \# instances avail.\ denotes the total number of complex-simple sentence pairs available for each corpus in the MultiSim benchmark. 
    \# eval.\ instances denotes the number of items sampled from all test, validation, and train splits (where necessary) for language used to evaluate performance on sentence simplification.
    }
    \label{tab:multisim_dataset_statistics}
\end{table*}

\section{Direct vs.\ Translated Evaluation for Non-English Dialogue Responses}
\label{app:direct_vs_translated_eval}

\begin{figure*}[ht]
    \centering
    \includegraphics[width=\textwidth]{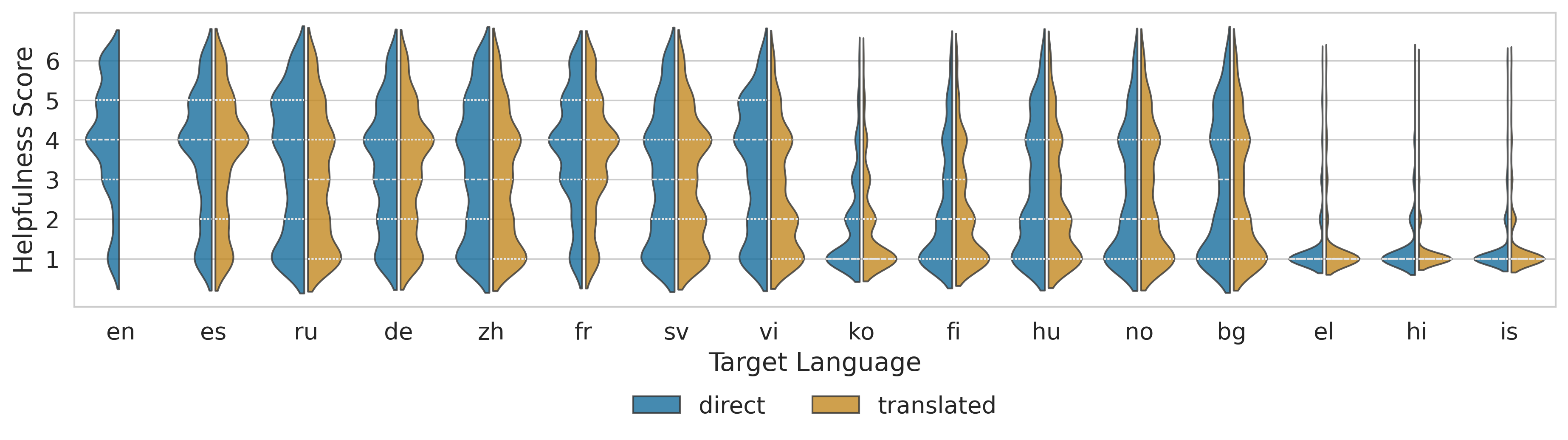}
    \caption{Distribution of helpfulness scores assigned by our LLM judge, GPT-3.5-Turbo, using direct evaluation in the specified target language and after translating target-language responses to English.}
    \label{fig:chat_llm_judge_llama_7b_translated_vs_direct_eval}
\end{figure*}

While using a powerful LLM to evaluate the outputs of other models has been shown to achieve reasonable agreement with human judgements in English \cite{zheng_judging_2023, chiang-lee-2023-large}, it is unclear whether this agreement transfers to all languages under investigation.
Recent work by \citet{hada-etal-2024-large} has shown that agreement between human and LLM judges tends to be lower for non-English languages, especially in the case of low-resource and non-Latin scripted languages, where the LLM judge tends to be overly optimistic in its assessment.
However, for certain assessment criteria, such as linguistic acceptability and general content quality, they also confirm that inter-annotator agreement between LLM-based evaluators and humans is in line with that of multiple human annotators.

To investigate this potential bias, we compared scores assigned by the  LLM judge on model-generated responses \textit{directly} in each non-English target language and on their \textit{English translations}. 
For each non-English prompt-response pair, we translate it into English using GPT-3.5-Turbo (\texttt{gpt-3.5-turbo-1106}) and the prompt shown in Figure \ref{fig:translating_alpacaeval_prompt}.
We pair the resulting translated responses with their corresponding English AlpacaEval prompts and repeat our LLM judge evaluation.

Figure \ref{fig:chat_llm_judge_llama_7b_translated_vs_direct_eval} shows that the distribution of assigned scores in the direct and translated evaluation settings is very similar for most languages. 
For languages that use non-Latin scripts (e.g., Chinese, Russian, Korean, etc.), we observe that the LLM judge tends to assign slightly higher scores more frequently when evaluating directly on the non-English prompt/response pairs.
This finding agrees with those from \citet{hada-etal-2024-large} and indicates that LLM-based evaluations in non-English languages can be overly optimistic and should be considered with caution.
Nevertheless, we observe that the discrepancy between direct and translated evaluation is relatively minor for the languages considered and leads to negligible differences in the overall average helpfulness score.
Based on these results, we opt to use the direct evaluation strategy for our experiments, which has the added benefit of avoiding the introduction of potential translation errors and keeping the cost of evaluation to a minimum.

\section{Results with Falcon 7b}
\label{app:falcon_results}

In order to assess whether our findings generalise to other LLMs, we repeat our experiments using Falcon 7b \cite{almazrouei2023falcon}.

\subsection{Single-turn Dialogue}
\label{sec:falcon_chat}

\begin{figure*}[ht]
    \centering
    \includegraphics[width=\textwidth]{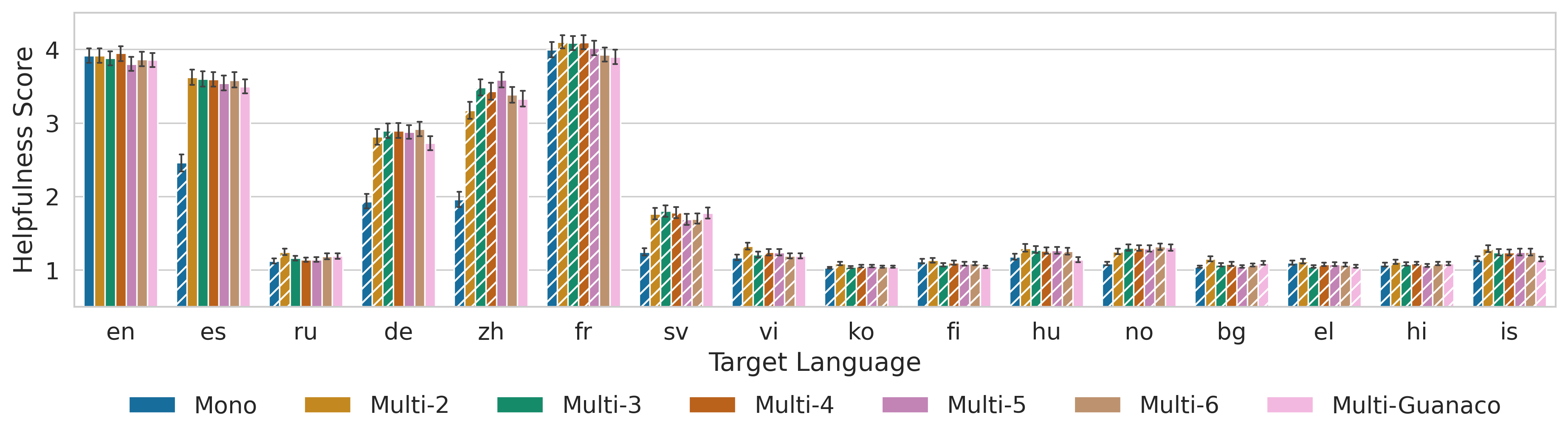}
    \caption{
    Average helpfulness of single-turn dialogue responses from Falcon 7b given incremental multilingual instruction tuning.
    Striped bars indicate a 0-shot setting and error bars show a confidence interval of 95\%.
    }
    \label{fig:chat_llm_judge_falcon_7b_incremental}
\end{figure*}

Figure \ref{fig:chat_llm_judge_falcon_7b_incremental} shows the helpfulness scores assigned by the LLM judge for Falcon 7b given incremental multilingual instruction tuning across all target languages.
Similarly to our results with Llama 2 7b (cf.\ Figure \ref{fig:chat_llm_judge_llama_7b_incremental}), cross-lingual transfer can be elicited after finetuning with relatively few languages, and no additional gains observed when including more than three languages.
Interestingly, Falcon 7b appears to show strong performance on French, even without multilingual finetuning, indicating that, despite being an English-centric model, it has strong capabilities in French out of the box.
For Spanish, German, and Chinese, performance is comparable to that of Llama 2 7b. 
However, for all other languages, responses are often ranked least helpful, indicating that Falcon 7b's multilingual capabilities are limited strictly to major European languages and Chinese.

\subsection{Sentence Simplification}
\label{sec:falcon_ts}

\begin{figure*}[ht]
    \centering
    \includegraphics[width=\textwidth]{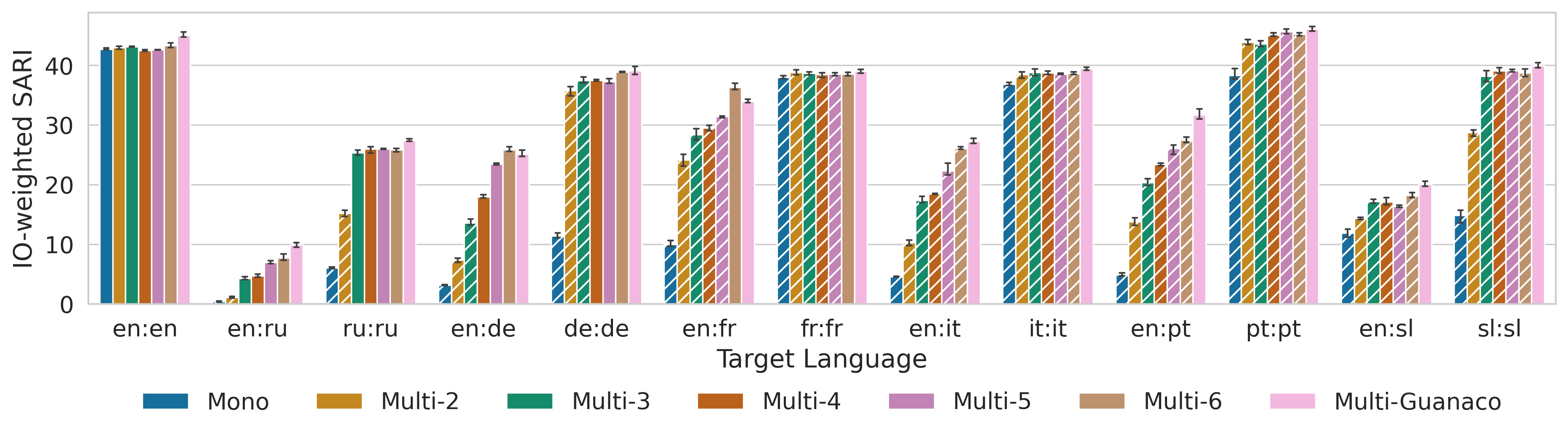}
    \caption{
    SARI weighted by IO language agreement for sentence simplification with Falcon 7b given incremental multilingual instruction tuning.
    Results are shown for both cross-lingual prompting (en:xx) and monolingual prompting (xx:xx)
    Striped bars indicate a 0-shot setting and error bars show a confidence interval of 95\%.
    }
    \label{fig:multisim_falcon_7b_incremental}
\end{figure*}

Figure \ref{fig:multisim_falcon_7b_incremental} depicts Falcon 7b's performance on the sentence simplification task given multilingual instruction tuning.
In general, the results emulate those seen with Llama 2 7b.
Performance on English remains uniform as multilinguality increases, while for non-English target languages, the largest jump in performance happens when moving from monolingual to bilingual instruction tuning, and plateauing again with just three languages.
The notable exceptions here are French and Italian, for which Falcon 7b appears to provide greater support even in the monolingual instruction tuning setting, echoing the results from the single-turn dialogue task.

\subsection{Extractive Question Answering}
\label{sec:falcon_xquad}

\begin{figure*}[ht]
    \centering
    \includegraphics[width=\textwidth]{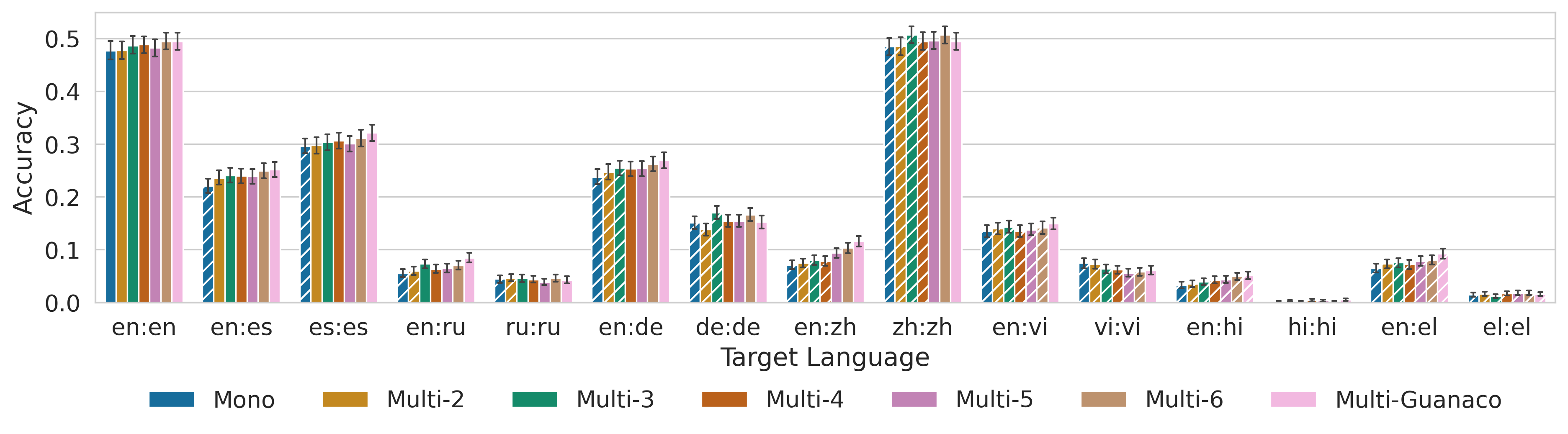}
    \caption{
    XQuAD results for Falcon 7b given incremental multilingual instruction tuning.
    Results are shown for both cross-lingual prompting (en:xx) and monolingual prompting (xx:xx). 
    Striped bars indicate a 0-shot setting and error bars show a confidence interval of 95\%.
    }
    \label{fig:xquad_falcon_7b_incremental}
\end{figure*}

Figure \ref{fig:xquad_falcon_7b_incremental} shows the results of Falcon 7b on XQuAD. 
While performance is generally lower than that of Llama 2 7b on this task (cf.\ Figure \ref{fig:xquad_llama_7b_incremental}), we observe a similar, albeit weaker, effect of multilingual finetuning within the supported languages (es, de, zh).

\subsection{Commonsense Reasoning}
\label{sec:falcon_xcsqa}

\begin{figure*}[ht]
    \centering
    \includegraphics[width=\textwidth]{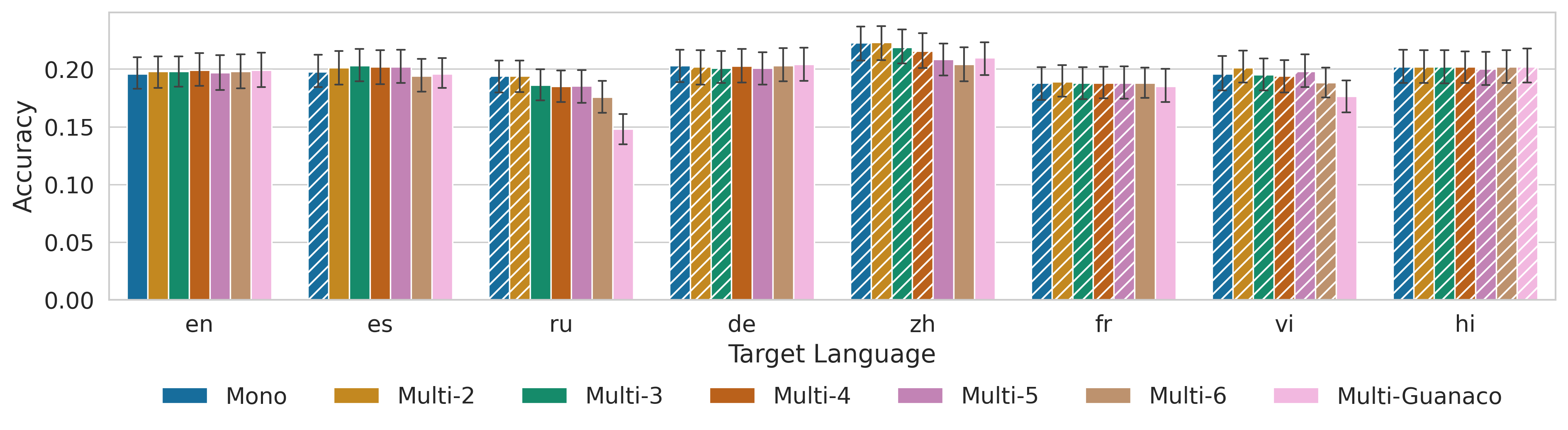}
    \caption{
    X-CSQA results for Falcon 7b given incremental multilingual instruction tuning.
    Striped bars indicate a 0-shot setting and error bars show a confidence interval of 95\%.
    }
    \label{fig:xcsqa_falcon_7b_incremental}
\end{figure*}

Figure \ref{fig:xcsqa_falcon_7b_incremental} shows the results of Falcon 7b on X-CSQA.
Strikingly, in contrast to the results achieved with Llama 2 7b (cf.\ Figure \ref{fig:xcsqa_llama_7b_incremental}), Falcon 7b fails to score above random performance across all target languages.
Regarding the effect of multilingual instruction tuning, we again see that it fails to deliver any performance improvements on this highly structured task.
Degradation on Russian and Vietnamese, is likely due to the model consistently failing to follow instructions and reproducing part of the question as the response.

\subsection{XNLI}
\label{sec:falcon_xnli}

\begin{figure*}[ht!]
    \centering
    \includegraphics[width=\textwidth]{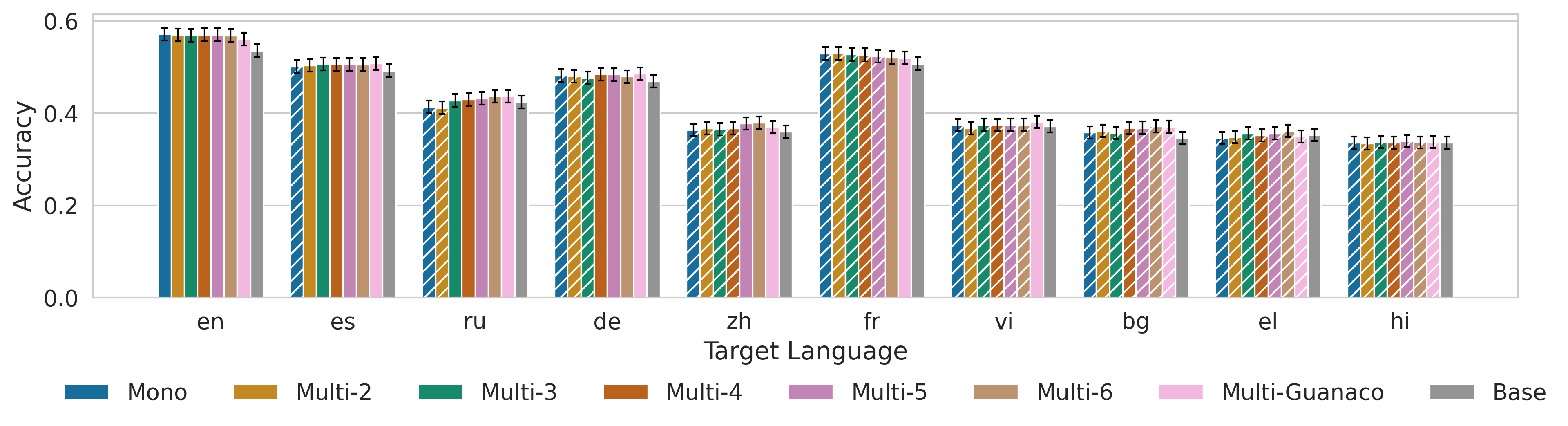}
    \caption{XNLI results for Falcon 7b given incremental multilingual instruction tuning.
    Striped bars indicate a 0-shot setting and error bars show a confidence interval of 95\%.
    }
    \label{fig:xnli_falcon_7b_incremental}
\end{figure*}

Figure \ref{fig:xnli_falcon_7b_incremental} shows the results of Falcon 7b on XNLI.
Similar to the results attained with Llama 2 7b (cf.\ Figure \ref{fig:xnli_llama2_7b_incremental}), we observe no significant differences in performance given different degrees of multilingual instruction tuning.

\subsection{Discussion}
\label{sec:falcon_discussion}

These additional experiments using Falcon 7b provide further support for our main findings: instruction tuning with as few as two languages significantly improves cross-lingual generalisation, enhancing performance in open-ended single-turn dialogue for some non-English languages.
However, strikingly, multilingual capabilities of Falcon are considerably narrower than that of Llama 2. 
This is also suggested by the statistics provided in Table \ref{tab:language_distributions}, which show a higher concentration of English and much lower proportions of non-English languages compared with both Llama 2 and GPT-3.
We suspect that this may be a result of stringent filtering of web-scraped pretraining data performed in producing the RefinedWeb corpus used to train the Falcon models \cite{penedo2023refinedweb}, which not only reduces the risk of potentially accidental contamination \cite{blevins-zettlemoyer-2022-language} but also language coverage.
That said, the RefinedWeb corpus comprises only part of Falcon's actual training corpus. 
\cite{almazrouei2023falcon} note that the final corpus also contains additional data drawn from curated sources including a European-focused (multilingual) Common-Crawl dataset, which could explain the strong performance on some major European languages.
In light of this, we suspect that additional Chinese data is also included in Falcon's training corpus, allowing for relatively strong performance in Chinese on the single-turn dialogue task.

\section{Results with Llama 3 8b}
\label{app:llama3_results}

In \S\ref{sec:scaling_experiments} we observed that scaling the base model from 7b to 70b parameters reduces the gap in performance between English and non-English target languages with regard to the single-turn dialogue task.
In this section, we evaluate Llama 3 8b \cite{metaai_introducing_2024}, which builds on Llama 2 by scaling up the pretraining data from 2 trillion tokens to 15 trillion tokens.
In Addition to a much larger pretraining corpus, Llama 3 is trained with a vocabulary of 128k tokens (vs.\ 32k for Llama 2), bringing it much closer to traditional multilingual models which typically contain roughly 250k tokens \cite{scao2023bloom, xue-etal-2021-mt5, conneau-etal-2020-unsupervised}.
While the exact language distribution of the model's pretraining data is not publicly known, it is reasonable to expect that these enhancements could lead to improvements in the model's non-English capabilities.

\begin{figure*}[ht]
    \centering
    \includegraphics[width=\textwidth]{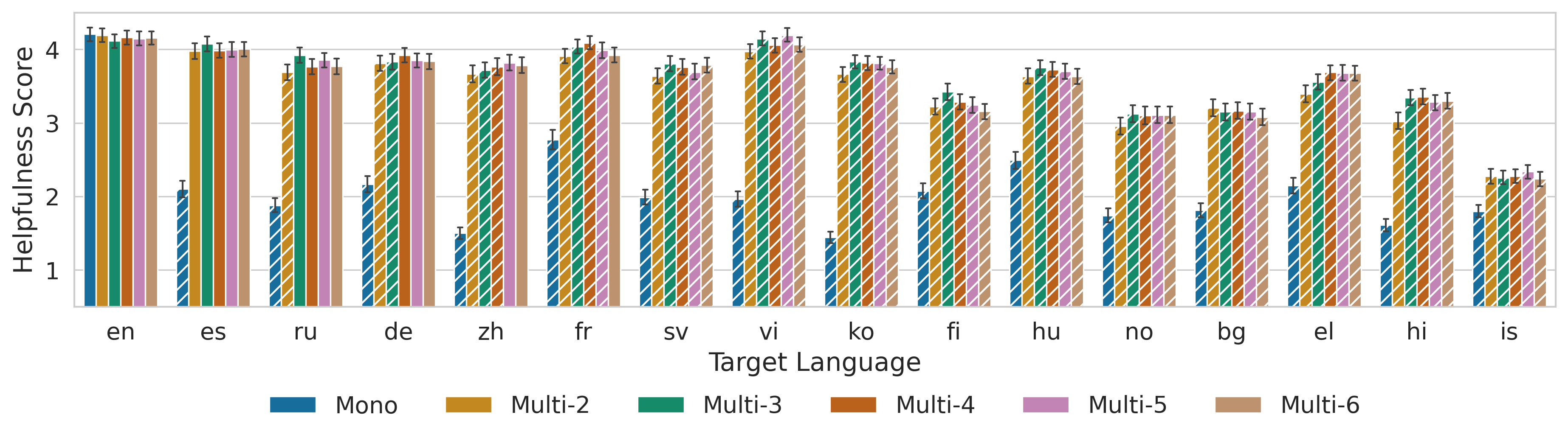}
    \caption{
    Single-turn dialogue performance of Llama 3 8b given incremental multilingual instruction tuning. 
    Striped bars indicate a 0-shot setting and error bars show a confidence interval of 95\%.
    }
    \label{fig:chat_llm_judge_llama3_incremental}
\end{figure*}

Figure \ref{fig:chat_llm_judge_llama3_incremental} shows the performance of Llama 3 8b, which closely resembles that of Llama 2 70b across most non-English languages, even outperforming it in languages like Vietnamese, Greek, and Hindi.
Most notably, we observe that our main findings still hold: multilingual instruction tuning with as few as two languages is both necessary and sufficient to elicit cross-lingual generalisation. 
Again, performance typically plateaus thereafter.
Given the improved performance on languages like Greek and Hindi that are low-resource under Llama 2's pretraining data distribution, we suspect that their representation in the pretraining data for Llama 3 is substantially higher. 
Still, performance on Icelandic suggests that this language may remain in the low-resource category under this model.

\section{Ablation Experiments}
\label{sec:ablation_experiments}

In Section \ref{sec:finetuning_data}, we constructed Multi-$i$ instruction tuning datasets of the same size by replacing a fixed number of English examples with examples from a new language. 
As a result, the proportion of non-English examples used for finetuning increases along with the number of languages. 
Table \ref{tab:multi_i_dataset_distributions} shows the exact makeup of each these datasets and how they relate to one another.
To rule out potential confounding factors between the proportion of non-English examples and the diversity of languages, we conduct ablation experiments in which we keep one variable fixed while incrementing the other.

\begin{table}[htb]
    \centering
    \small
    \begin{tabular}{llrr}
        Dataset & Languages & \% NE & Total \\ 
        \cmidrule(lr){1-1} \cmidrule(lr){2-2} \cmidrule(lr){3-3} \cmidrule(lr){4-4}
        \addlinespace
        Mono & en & 0.00\% & 3,200 \\
        Multi-2 & en, es  & 6.25\% & 3,200 \\
        Multi-3 & en, es, ru  & 12.50\% & 3,200 \\
        Multi-4 & en, es, ru, de  & 18.75\% & 3,200 \\
        Multi-5 & en, es, ru, de, zh  & 25.00\% & 3,200 \\
        Multi-6 & en, es, ru, de, zh, fr  & 31.25\% & 3,200 \\       
        \lightbottomrule
    \end{tabular}
    \caption{
    Makeup of the incremental multilingual instruction-tuning datasets. 
    }
    \label{tab:multi_i_dataset_distributions}
\end{table}

\begin{figure*}[htb]
    \centering
    \begin{subfigure}{.45\textwidth}
    \includegraphics[width=\linewidth]{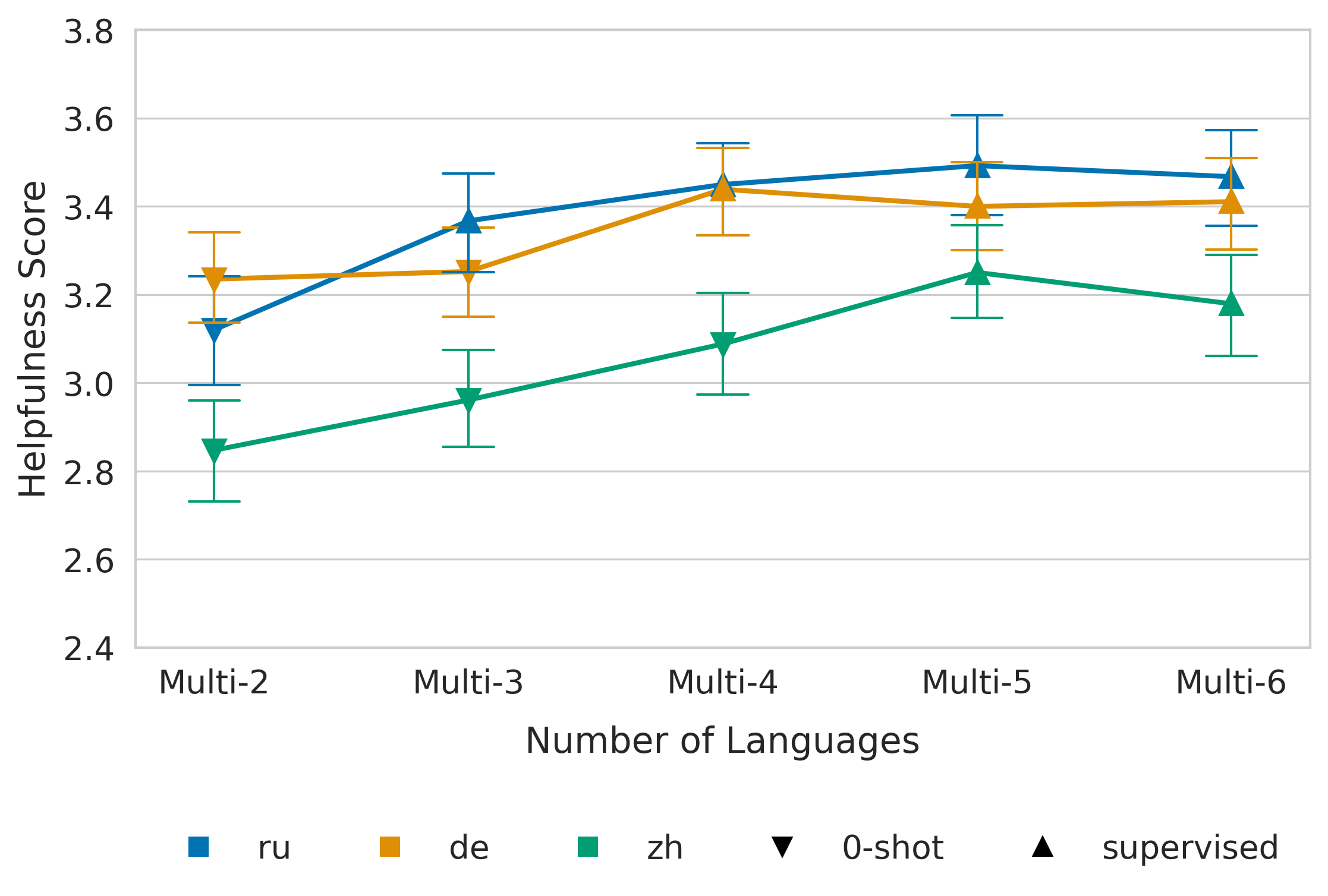}
    \caption{Impact of incrementing the number of languages while keeping the total number of non-English examples fixed at 400 (12.5\% of the finetuning data).
    0-shot and supervised settings are distinguished using $\blacktriangledown$ and $\blacktriangleup$, respectively.
    \newline
    }
    \label{fig:chat_llm_judge_llama_7b_incremental_languages_n400}
    \end{subfigure}
    \hspace{1em}
    \begin{subfigure}{.45\textwidth}
    \centering
    \includegraphics[width=\linewidth]{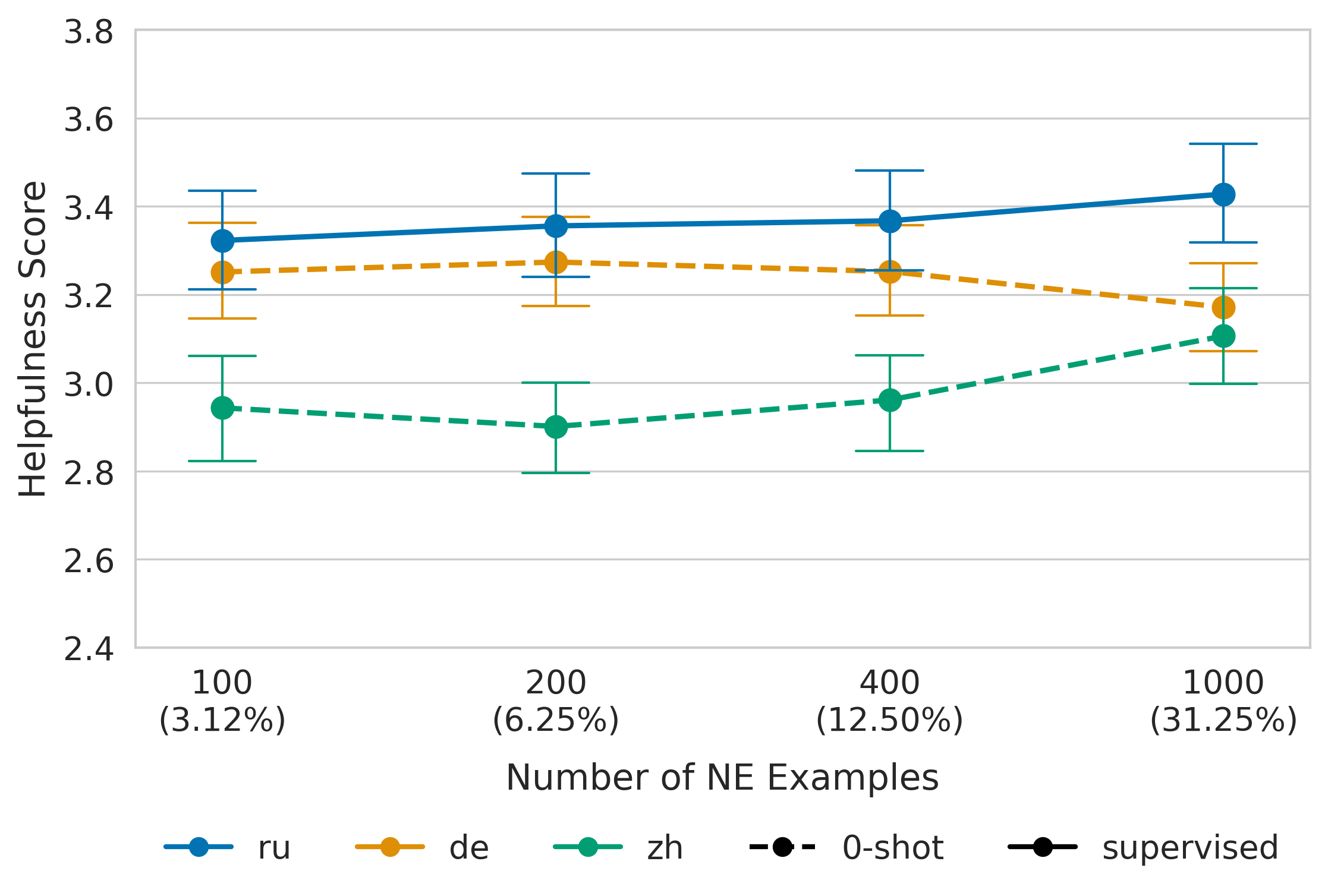}
    \caption{Impact of incrementing the total number of non-English examples while keeping the number of languages fixed at three (English, Spanish, and Russian). 
    Here, Russian is seen during finetuning, while German and Chinese are in the 0-shot setting.
    }
    \label{fig:chat_llm_judge_llama_7b_incremental_examples_ml2}
    \end{subfigure}
    \caption{Average helpfulness of single-turn dialogue responses from Llama 2 7b given multilingual instruction tuning on a fixed budget (limiting either the number of non-English instances \ref{fig:chat_llm_judge_llama_7b_incremental_languages_n400} or the number of languages \ref{fig:chat_llm_judge_llama_7b_incremental_examples_ml2}).
    Error bars show a confidence interval of 95\%.
    }
\end{figure*}

Figure \ref{fig:chat_llm_judge_llama_7b_incremental_languages_n400} shows that when using a fixed budget of 400 non-English examples (which equates to 12.5\% of the finetuning data) and incrementing only the number of languages within this budget, performance increases consistently as each new language is added, closely reflecting the results of our main experiments.
In contrast, Figure \ref{fig:chat_llm_judge_llama_7b_incremental_examples_ml2} shows that when training with a fixed number languages (e.g., Multi-2) and incrementing the number of non-English examples (in this case, Spanish), performance tends to remain lower and generally shows less improvement.
These results underscore our main findings discussed in \S \ref{sec:chat} and \S \ref{sec:discussion}.

\begin{figure*}[ht]
  \centering
  \begin{subfigure}{\textwidth}
    \includegraphics[width=\textwidth]{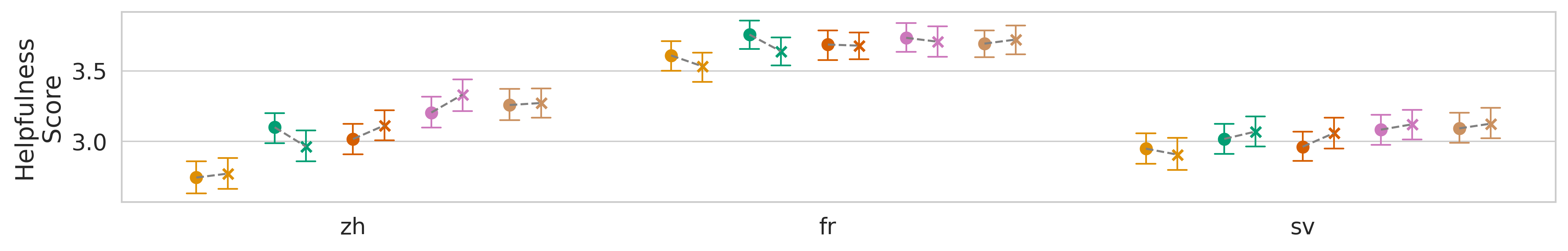}
  \end{subfigure}
  \\
  \begin{subfigure}{\textwidth}
    \includegraphics[width=\textwidth]{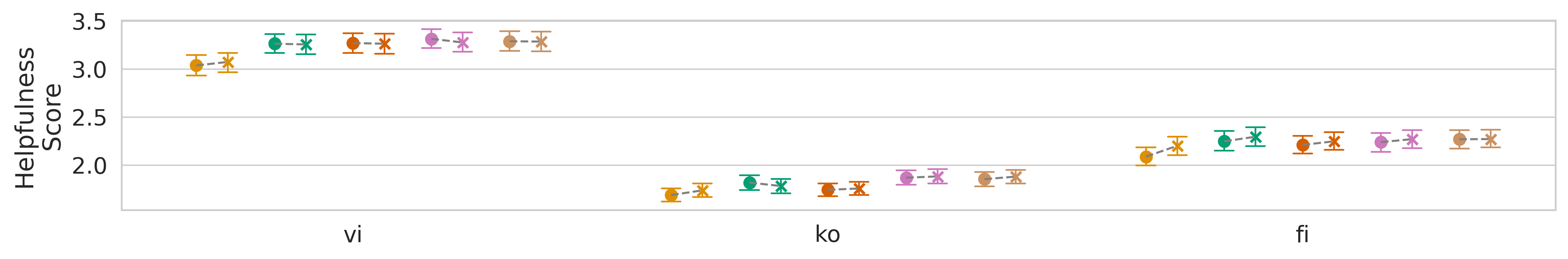}
  \end{subfigure}
  \\
  \begin{subfigure}{\textwidth}
    \includegraphics[width=\textwidth]{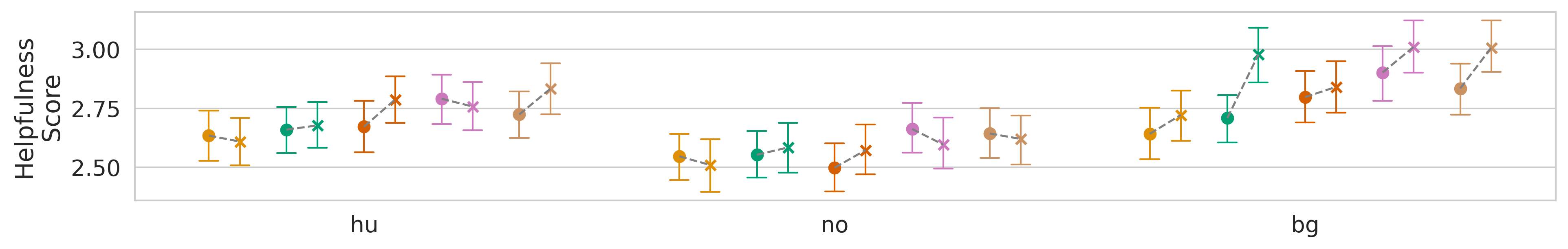}
  \end{subfigure}
  \\
  \begin{subfigure}{\textwidth}
    \includegraphics[width=\textwidth]{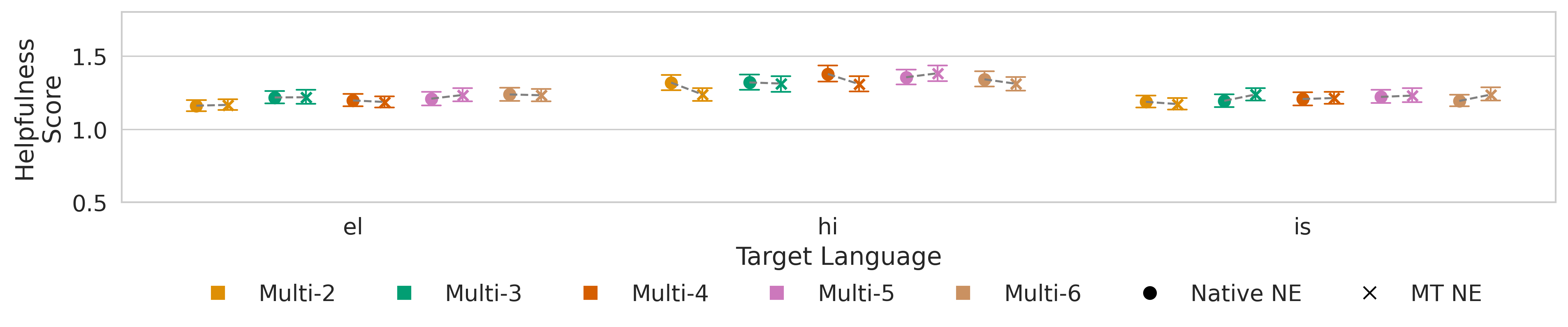}
  \end{subfigure}
  \caption{
  Comparison of Llama 2 7b with multilingual instruction tuning using native (Native NE) and translated non-English (MT NE) examples for additional target languages not shown in Figure \ref{fig:chat_llm_judge_llama_7b_incremental_diff_mt}.
  Error bars show a confidence interval of 95\%.
  }
\label{fig:chat_llm_judge_llama_7b_incremental_diff_mt_additional_langs}
\end{figure*}

\begin{figure*}[ht!]
\centering
\begin{tikzpicture}[scale=0.8, every node/.style={transform shape}]
\node[rectangle, rounded corners, draw=blue!20, fill=blue!10, text width=\textwidth, align=left, inner sep=1.2ex, font=\ttfamily] (prompt) {
You are an expert language evaluator. 
~\\
~\\
You are evaluating a response that has been submitted for a particular task, using a specific set of standards. Below is the data:
~\\
~\\
{[BEGIN DATA]}
~\\
{***}
~\\
{[Task]}: \{\{task\}\}
~\\
{***}
~\\
{[Submission]}: \{\{submission\}\}
~\\
{***}
~\\
{[Criterion]}: helpfulness:
~\\
``1'': ``Not helpful - The generated text is completely irrelevant, unclear, or incomplete. It does not provide any useful information to the user.''
~\\
``2'': ``Somewhat helpful - The generated text has some relevance to the user’s question, but it may be unclear or incomplete. It provides only partial information, or the information provided may not be useful for the user’s needs.''
~\\
``3'': ``Moderately helpful - The generated text is relevant to the user’s question, and it provides a clear and complete answer. However, it may lack detail or explanation that would be helpful for the user.''
~\\
``4'': ``Helpful - The generated text is quite relevant to the user’s question, and it provides a clear, complete, and detailed answer. It offers additional information or explanations that are useful for the user. However, some of the points of the response are somewhat repetitive or could be combined for greater clarity and concision.''
~\\
``5'': ``Very helpful - The generated text is highly relevant to the user’s question, and it provides a clear, complete, and detailed answer. It offers additional information, explanations, or analogies that are not only useful but also insightful and valuable to the user. However, the structured of the response is not well-organized and there is no clear progression or logical sequence of different points in the response.''
~\\
``6'': ``Highly helpful - The generated text provides a clear, complete, and detailed answer. It offers additional information or explanations that are not only useful but also insightful and valuable to the user. The response is also in a logical and easy-to-follow manner by explicitly using headings, bullet points, or numbered lists to break up the information and make it easier to read.''
~\\
{***}
~\\
{[END DATA]}
~\\
~\\
Does the submission meet the criterion? First, write out in a step by step manner your reasoning about the criterion to be sure that your conclusion is correct. Avoid simply stating the correct answers at the outset. 
Provide your response as a RFC8259 compliant JSON following this schema:
~\\
~\\
\{``reasoning'': str, ``score'': int\}
};

\end{tikzpicture}
\caption{Prompt template used to query the LLM judge in order to assess the helpfulness of a single-turn dialogue response. This prompt closely follows the one used by \citet{zhou2023lima} but extends it by specifying the return value as a valid JSON-like object.}
\label{fig:eval_helpfulness_prompt}
\end{figure*}

\begin{table*}[ht]
    \centering
    \small
    \begin{tabularx}{0.8\textwidth}{lX}
    \lighttoprule
    Prompt: &  \texttt{\#\#\# Human: Please rewrite the following complex sentence in order to make it easier to understand.
    You can do so by replacing complex words with simpler synonyms (i.e.\ paraphrasing), deleting unimportant information (i.e.\ compression), and/or splitting a long complex sentence into several simpler ones.
    The final simplified sentence needs to be grammatical, fluent, and retain the main ideas of its original counterpart without altering its meaning.}
    \newline
    \newline
    \texttt{Complex sentence:}
    Sie gelten als eiskalte Fress-Maschinen.
    \newline
    \newline
    \texttt{\#\#\# Assistant:}
    \\
    \lightmidrule
    Reference: & Sie gelten als Fress-Maschinen.
    \\
    \lightmidrule
    System: & They are seen as cold-blooded eaters.
    \\
    \lightbottomrule
    \end{tabularx}
    \caption{
    An example of the 0-shot sentence simplification task. 
    Monospaced font represents the prompt template used, adapted from instructions provided to crowdworkers for the creation of the ASSET dataset \cite{alva-manchego-etal-2020-asset}.
    This example shows the cross-lingual prompt strategy (en:xx), where the task instruction is provided in English but the source sentence is in German.
    Ideally, the system language should match the source sentence, which it fails to do in this case.
    }
    \label{tab:ts_prompt}
\end{table*}

\begin{table*}[ht]
    \centering
    \small
    \begin{tabularx}{0.8\textwidth}{lX}
    \lighttoprule
    Prompt: & \texttt{\#\#\# Human: Answer the question from the given passage. Your answer should be directly extracted from the passage, and it should be a single entity, name, or number, not a sentence.}
    \newline
    \newline
    \texttt{Passage:}
    \newline
    \newline
    Die Verteidigung der Panthers gab nur 308 Punkte ab und belegte den sechsten Platz in der Liga, während sie die NFL mit 24 Interceptions in dieser Kategorie anführte und sich mit vier Pro Bowl-Selektionen rühmen konnte. Pro Bowl Defensive Tackle Kawann Short führte das Team mit 11 Sacks an, erzwang zudem drei Fumbles und erzielte zwei Fumble Recoverys. Mario Addison, ebenfalls Lineman, addierte $6\frac{1}{2}$ Sacks hinzu. Die Panthers-Line präsentierte auch den erfahrenen Defensive End Jared Allen, einen 5-fachen Pro-Bowler, der mit 136 Sacks der aktive Anführer in der NFL-Kategorie Karriere-Sacks war, sowie den Defensive End Kony Ealy, der 5 Sacks in nur 9 Starts erzielte. Nach ihnen wurden zwei der drei Linebacker der Panthers ausgewählt, um im Pro Bowl zu spielen: Thomas Davis und Luke Kuechly. Davis erzielte $5\frac{1}{2}$ Sacks, vier erzwungene Fumbles und vier Interceptions, während Kuechly das Team bei den Tackles anführte (118), zwei Fumbles erzwang und vier Pässe abfing. Carolinas Secondarys bestanden aus dem Pro Bowl-Safety Kurt Coleman, der das Team mit einem Karrierehoch von sieben Interceptions anführte und gleichzeitig 88 Tackles erzielen konnte, und Pro Bowl-Cornerback Josh Norman, der sich während der Saison zur Shutdown Corner entwickelte und vier Interceptions erzielte, von denen zwei zu Touchdowns für sein Team wurden.
    \newline
    \newline
    \texttt{Question:}
    \newline
    \newline
    Mit wie vielen Interceptions erzielte Josh Norman im Jahr 2015 Touchdowns?
    \newline
    \newline
    \texttt{Note: Your answer should be directly extracted from the passage and be a single entity, name, or number. Provide the answer in quotations.}
    \newline
    \newline
    \texttt{\#\#\# Assistant: Based on the passage, the answer to the question is ``} \\
    \lightmidrule
    Reference: & zwei \\
    \lightmidrule
    System: & two".
    \newline
    \newline
    In the passage, it is stated that Josh Norman ``erzielte vier Interceptions, von denen zwei zu Touchdowns für sein Team wurden.'' This means that Norman intercepted four passes during the season, and two of those interceptions were returned for touchdowns. \\
    \lightbottomrule
    \end{tabularx}
    \caption{
    Example input prompt, reference, and model response for the 0-shot extractive QA task using XQuAD. Monospaced font represents the prompt template used for each test instance.
    Here, we show the cross-lingual prompting strategy (en:xx), where the task instruction and response prefix is provided English, while the context passage and question are in the target language (in this case, German).
    A correct answer in this setting requires the model to continue the response by switching into the target language, which it fails to do in this particular example.
    }
    \label{tab:xquad_example}
\end{table*}

\begin{table*}[ht]
    \centering
    \small
    \begin{tabularx}{0.8\textwidth}{lX}
    \lighttoprule
    Prompt: &  
    \texttt{\#\#\# Human: Dir wird eine Frage vorgelegt, die mehrere Antwortmöglichkeiten hat. Wähle die am besten geeignete Option aus ``A'', ``B'', ``C'', ``D'' oder ``E'' basierend auf deinem gesunden Menschenverstand. Bitte gebe deine Antwort in Form eines einzelnen Buchstabens in Anführungszeichen an.}
    \newline
    \newline
    \texttt{Frage:}
    \newline
    \newline
    Sarah stellte ihr Essen in einen Mülleimer neben ihrem Schreibtisch.  Wo mag sie gewesen sein?
    \newline
    \newline
    \texttt{Antwortmöglichkeiten:}
    \newline
    \newline
    \texttt{A:} Bushaltestelle
    \newline
    \texttt{B:} Ecke
    \newline
    \texttt{C:} Motel
    \newline
    \texttt{D:} Park
    \newline
    \texttt{E:} Büro
    \newline
    \newline
    \texttt{\#\#\# Assistant: ``}
    \\
    \lightmidrule
    Reference: & E \\
    \lightmidrule
    System: & E'' \\
    \lightbottomrule
    \end{tabularx}
    \caption{
    Example of the commonsense reasoning task using X-CSQA. Monospaced font represents the prompt template used for each test instance.
    Here, the prompt strategy is monolingual (xx:xx) with the task instruction provided in the same language as the test instance (in this case, German).
    A correct answer requires the model to output the letter corresponding to the ground-truth answer.
    }
    \label{tab:xcsqa_example}
\end{table*}

\begin{table*}[htb]
    \center
    \small
    \begin{tabularx}{0.8\textwidth}{c X c}
    \lighttoprule
    \multicolumn{3}{c}{XNLI prompt template: \texttt{\{Sentence 1\}, right?\ \{label\}, \{Sentence 2\}}} \\
    \lightmidrule
    Label & \multicolumn{1}{c}{Prompt query} & NLL \\
    \lightmidrule
    \multirow{2}{*}{Entailment} & In measuring effectiveness, perfection is unattainable., right? \underline{Yes}, You can never be perfect. & -113.94 \\
    \multirow{2}{*}{Neutral} & In measuring effectiveness, perfection is unattainable., right? \underline{Also}, You can never be perfect. & -116.79 \\
    \multirow{2}{*}{Contradiction} & In measuring effectiveness, perfection is unattainable., right? \underline{No}, You can never be perfect. & -114.01 \\
    \lightbottomrule
    \end{tabularx}
    \caption{
        Example prompt queries for XNLI given the predefined template used in the LM Evaluation Harness \cite{gao-etal-2023-eval_harness} and the three possible labels.
        Underlined words are the language-specific connectors corresponding to the three possible labels.
        The final answer is the sequence with the highest negative log-likelihood (NLL) according to the model.
        In this case, Llama 2 7b Mono correctly identifies the relationship between the two sentences as entailment.
        }
    \label{tab:xnli_example}
\end{table*}

\end{document}